\documentclass[]{fairmeta}
\usepackage{microtype}
\usepackage{amsfonts}
\usepackage{graphicx}
\usepackage{booktabs}
\usepackage{wrapfig}
\usepackage{amsmath}
\usepackage{amsthm}
\usepackage{algpseudocode}
\usepackage[linesnumbered,lined,boxed,commentsnumbered,ruled,longend]{algorithm2e}
\usepackage[capitalize,noabbrev]{cleveref}
\theoremstyle{plain}
\newtheorem{theorem}{Theorem}[section]

\theoremstyle{definition}
\newtheorem{definition}[theorem]{Definition}

\theoremstyle{remark}

\usepackage{enumitem}
\newcommand{\Sys}{\textsc{MagicPIG}\xspace}
\newcommand{\sys}{\textsc{MagicPIG}\xspace}
\newcommand{\TopK}{{$\mathrm{TopK}$}\xspace}

\usepackage{amsmath,amsfonts,bm}

\def\eqref#1{equation~\ref{#1}}

\def\1{\bm{1}}

\def\eps{{\epsilon}}

\DeclareMathAlphabet{\mathsfit}{\encodingdefault}{\sfdefault}{m}{sl}
\SetMathAlphabet{\mathsfit}{bold}{\encodingdefault}{\sfdefault}{bx}{n}

\title{\includegraphics[width=0.04\linewidth]{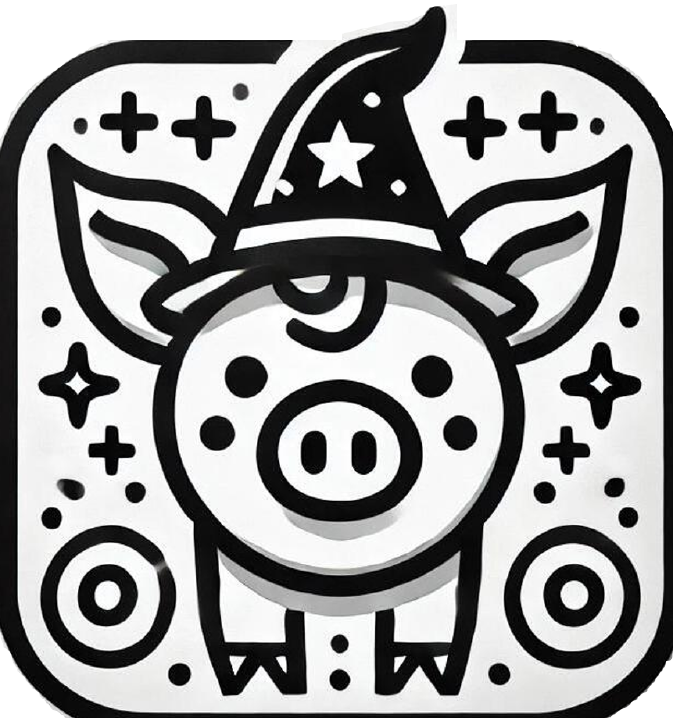} \textsc{MagicPIG}: LSH Sampling for Efficient LLM Generation}

\author[\dagger]{Zhuoming Chen}
\author[\dagger]{Ranajoy Sadhukhan}
\author[\ddagger]{Zihao Ye}
\author[\dagger]{Yang Zhou}
  \author[\S \sharp]{Jianyu Zhang}
  \author[\sharp]{Niklas Nolte}
  \author[\sharp]{Yuandong Tian}
  \author[\sharp]{ Matthijs Douze}
  \author[\S \sharp]{Leon Bottou}
  \author[\dagger]{Zhihao Jia}
  \author[\dagger]{Beidi Chen}

  \affiliation[\dagger]{Carnegie Mellon University}
  \affiliation[\ddagger]{University of Washington}
  \affiliation[\S]{New York University}
  \affiliation[\sharp]{Meta AI}

\abstract{
 Large language models (LLMs) with long context windows have gained significant attention. However, the KV cache, stored to avoid re-computation, becomes a bottleneck. Various dynamic sparse or TopK-based attention approximation methods have been proposed to leverage the common insight that attention is sparse. In this paper, we first show that TopK attention itself suffers from quality degradation in certain downstream tasks because attention is not always as sparse as expected. Rather than selecting the keys and values with the highest attention scores, sampling with theoretical guarantees can provide a better estimation for attention output. To make the sampling-based approximation practical in LLM generation, we propose \textsc{MagicPIG}, a heterogeneous system based on Locality Sensitive Hashing (LSH). \textsc{MagicPIG} significantly reduces the workload of attention computation while preserving high accuracy for diverse tasks. \textsc{MagicPIG} stores the LSH hash tables and runs the attention computation on the CPU, which allows it to serve longer contexts and larger batch sizes with high approximation accuracy. \textsc{MagicPIG} can improve decoding throughput by up to $5\times$ across various GPU hardware and achieve 54ms decoding latency on a single RTX 4090 for Llama-3.1-8B-Instruct model with a context of 96k tokens.
}

\metadata[Github]{\url{https://github.com/Infini-AI-Lab/MagicPIG}}
\metadata[Website]{\url{https://www.lsh-ai.com}}

\begin{document}

\maketitle
\section{Introduction}
\begin{wrapfigure}{r}{150px}
    \centering
        \vspace{-6mm}
    \includegraphics[width=135px]{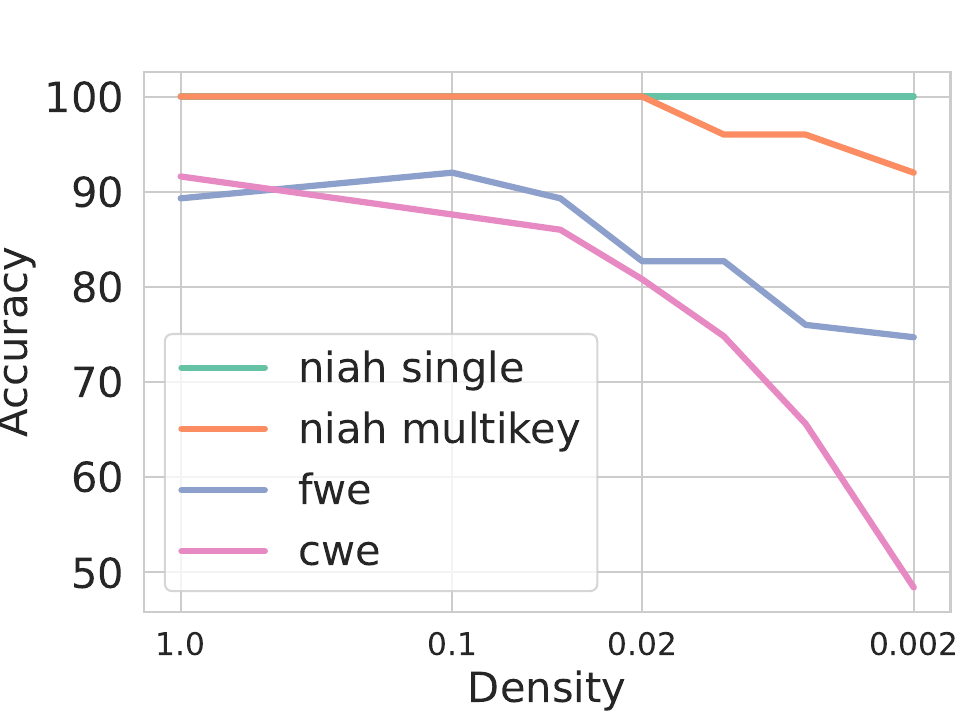}
    \vspace{-2mm}
    \caption{While \TopK attention performs well on retrieval tasks (niah) where the useful information reduces to a few words, it degrades severely in aggregated tasks like word extraction (cwe, fwe).
    x-axis: proportion of attention keys used for \TopK attention.
    }
    \label{fig: topkperf}
        \vspace{-3mm}
\end{wrapfigure}
Large language models (LLMs) with long context windows, such as GPT~\citep{achiam2023gpt}, Llama~\citep{dubey2024llama}, and Gemini~\citep{team2023gemini}, have gained significant attention for their ability to enhance applications like chatbots~\citep{chiang2024chatbot}, search engines~\citep{wang2024leave}, and video analysis~\citep{cheng2024videollama}. However, serving long-context LLMs is highly challenging due to the unique bottleneck in auto-regressive generation—the key-value (KV) cache, which stores intermediate attention keys and values to avoid re-computation~\citep{pope2022efficiently,NEURIPS2023_6ceefa7b}. 
Specifically, the KV cache grows linearly with both the batch size and sequence length, occupying substantial GPU memory and increasing decoding time. 
Moreover, the KV cache makes LLM generation extremely memory-bound, leading to underutilization of GPU computational power. 
For instance, an NVIDIA A100-40GB GPU can only handle a single request for Llama with a 128k context length, with nearly half of the decoding time spent accessing the KV cache, and poor GPU utilization ~\citep{he2024fastdecode}.

Leveraging the common insight that attention is naturally sparse, dynamic sparse or \TopK-based approximation has been extensively studied~\citep{tang2024quest,singhania2024loki,zhang2024pqcache,wu2024retrieval}, but three major challenges prevent a wide adoption in LLM serving systems. 
(1) \textbf{Quality Degradation.} They usually propose various strategies to approximate a subset of KV cache that yields the highest attention scores. However, \TopK attention itself is a biased attention approximation and lacks theoretical guarantees. \Cref{fig: topkperf} shows that even exact \TopK attention results significantly degrade the accuracy of certain downstream tasks. 
(2) \textbf{High Overhead.} There is a large overhead to identify \TopK attention, which becomes the bottleneck rather than the attention computation. For example, as studied in~\citet{liu2024retrievalattention}, naively applying a search algorithm like IVF~\citep{douze2024faiss} requires access over $30\%$ key states to obtain the exact \TopK, showing an unsatisfying trade-off between search accuracy and cost.
(3) \textbf{No Memory Saving.} Although saving KV cache loading time, they cannot reduce the total memory occupied by the KV cache, which limits the maximum context and batch sizes when VRAM is scarce.  

\begin{figure*}
    \centering
    \vspace{-5mm}
    \subfloat[Long tailed phenomena]{
\includegraphics[width=0.32\linewidth]{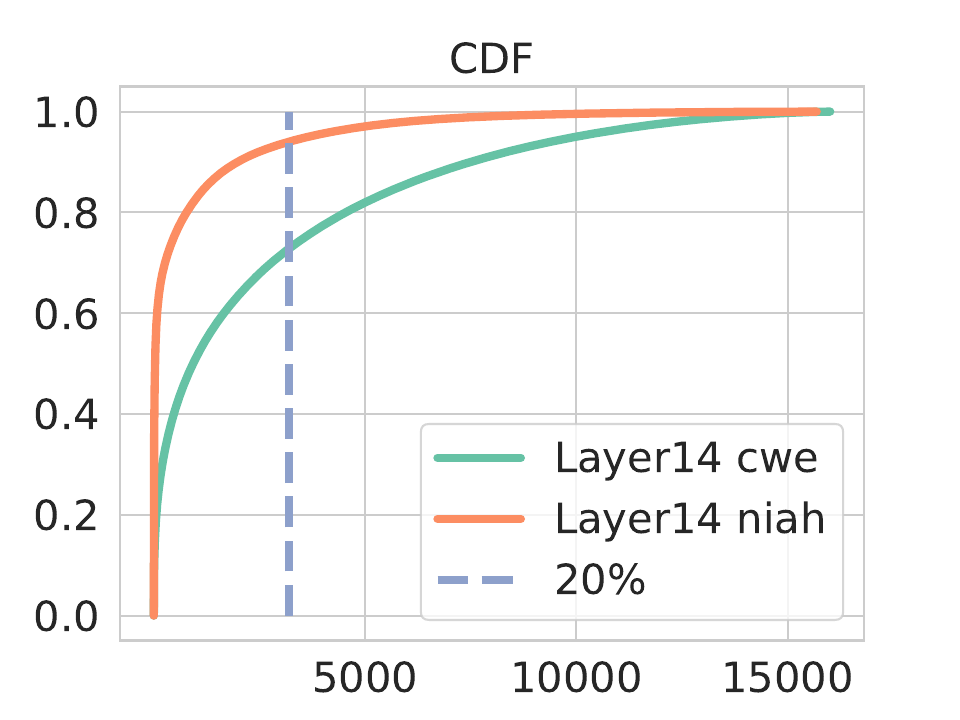}
    \label{fig: longtail}
    }
  \subfloat[Attention sink reshapes sparsity]{
\includegraphics[width=0.32\linewidth]{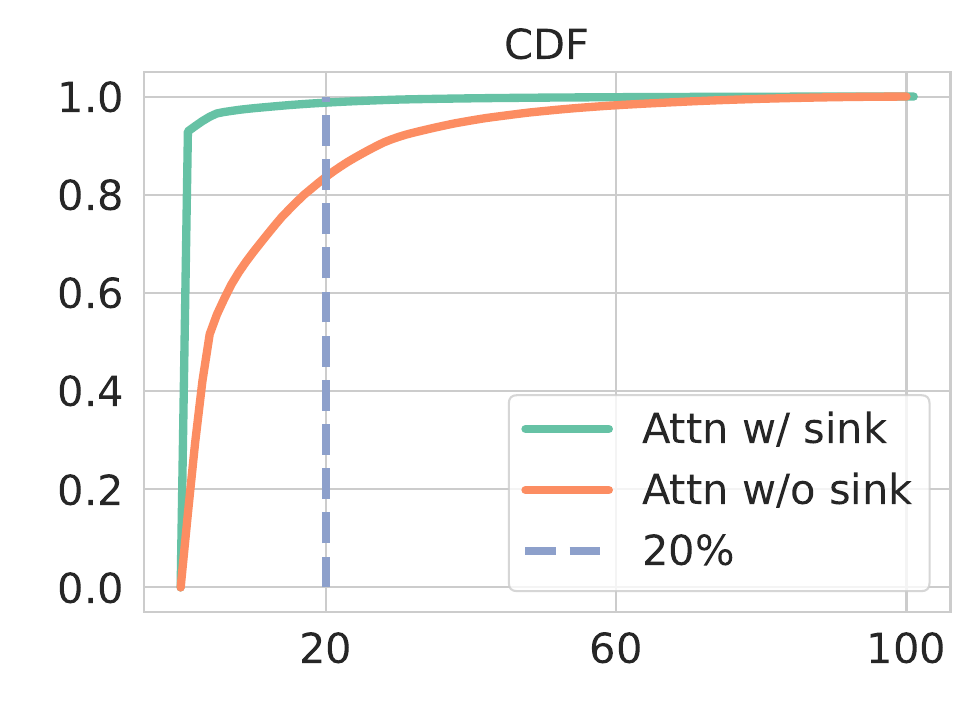}
    \label{fig: influenceofsink} 
    }
    \subfloat[Geometry of attention]{
\includegraphics[width=0.32\linewidth]{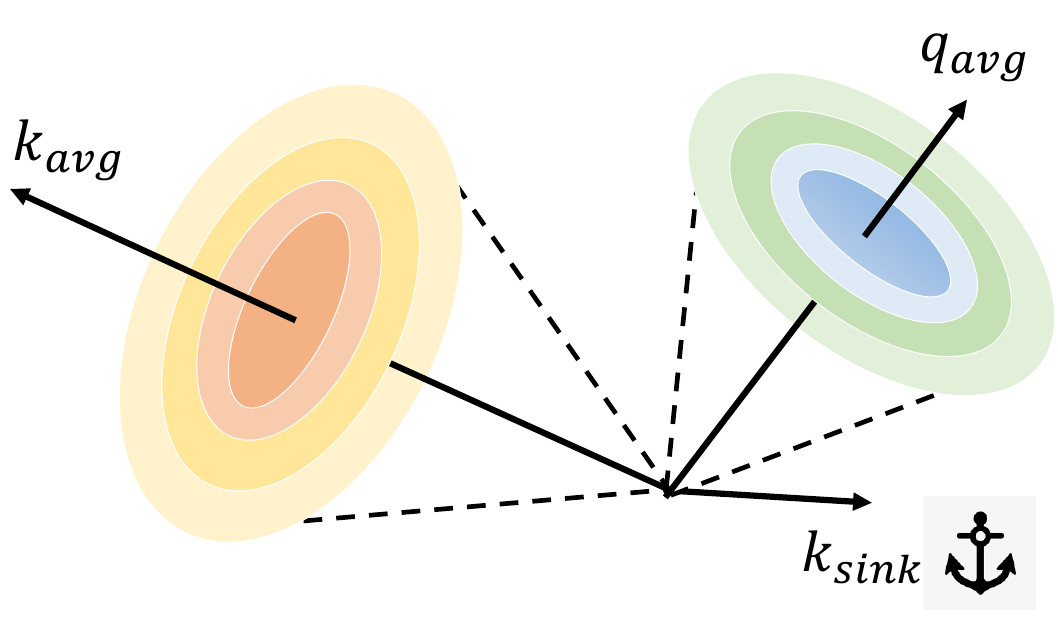}
    \label{fig: geometry} 
    }
    \caption{\textbf{Left:} Examples of long-tailed distribution in LLM. 
    The x-axis is the fraction (or number of tokens) used in the \TopK, a.k.a. the \emph{sampling budget}.
    \textbf{Mid:} Sink tokens make attention score look sparser. \textbf{Right:} The geometry of attention. The key of attention sink $k_{sink}$ is almost opposite to other tokens, and its orientation is surprisingly invariant with input tokens. Query states lie close to $k_0$, thus forming attention sink and~\Cref{fig: influenceofsink}. $k$ usually lies in a narrow cone that is far away from $q$. In certain heads, this geometry will result in a long-tailed distribution of attention score and difficulty searching for the \TopK keys.
    }
    \vspace{-5mm}
\end{figure*}

An ideal sparse attention approximation approach should (1) preserve full accuracy for a diverse set of downstream tasks with guarantees, (2) involve low-cost overhead for KV cache selection, and (3) save GPU memory. The following observations, together with the performance drop shown in~\Cref{fig: topkperf} suggest that to achieve such demanding requirements, we need to go beyond \TopK attention:
\begin{itemize}
    \item \textit{\underline{Attention is not always sparse}}. Contradictory to previous belief~\citep{NEURIPS2023_6ceefa7b,zhang2024pqcache,tang2024quest,liu2024retrievalattention}, we observe that attention is not always sparse, especially for tasks that leverage the full context. As shown in~\Cref{fig: longtail}, in some layers, attention distribution can be very long-tailed, \textit{i.e.}, the $\mathrm{Top} 20\%$ attention can only cover $70\%$ of the total attention scores.
    \item  \textit{\underline{Seemingly high sparsity is usually a consequence of an attention sink.}} Most of the attention scores concentrate on initial tokens (attention sink phenomenon)~\citep{xiao2023efficient}, making the distribution look sparser. However, as shown in~\Cref{fig: influenceofsink}, attention scores are distributed more uniformly among tokens except for the sink. According to the geometrical interpretation of sink, keys, and queries shown in ~\cref{fig: geometry}, the attention sink, which we found surprisingly almost static regardless of the input token, is just for imposing sparsity on the attention distribution.
    \item \textit{\underline{It is hard to find \TopK attention.}} \cref{fig: geometry} also shows why searching for the Top-K keys is intrinsically costly. The keys and queries usually lie within two narrow cones with nearly opposite orientations, except for the attention sink. This significant mismatch between query and data distributions causes nearest-neighbor search methods to perform poorly.
    \end{itemize}

These limitations of \TopK attention require rethinking the sparse attention approximation. 
Rather than only using the keys and values with the highest scores, leveraging information on the distribution can make the estimation more accurate. 
We approach this as a bias correction problem in sampling.
Unbiased and efficient sampling has been long studied in biology~\citep{lukacs2009closed}, sociology~\citep{chen2018unique} as well as machine learning~\citep{NEURIPS2019_a2ce8f17,chen2019fast,zandieh2023kdeformer}, with theoretical guarantees. 

\begin{wrapfigure}{r}{140px}
    \centering
      \hspace{-2mm}\includegraphics[width=140px]{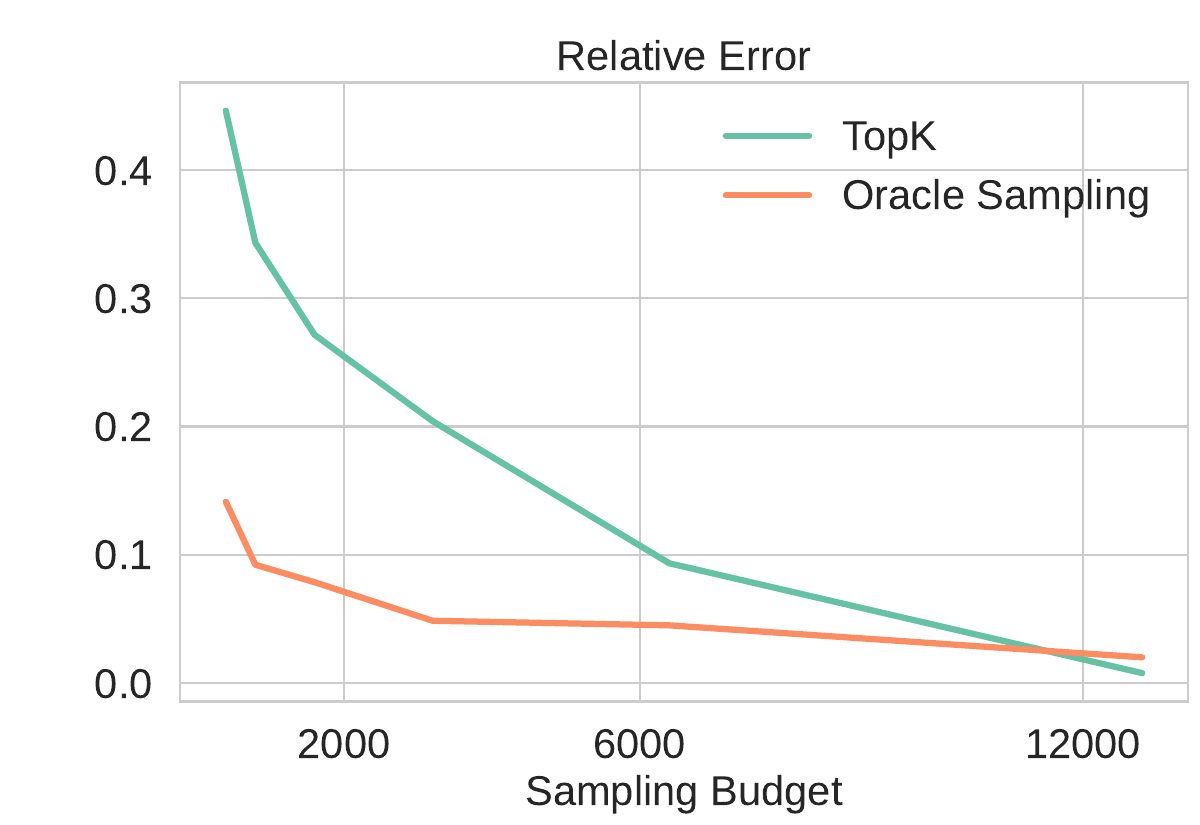}
        \vspace{-3mm}
    \caption{\TopK v.s. Sampling, 16k total context}
    \label{fig: topkvsos}
        \vspace{-4mm}
\end{wrapfigure}

~\Cref{fig: topkvsos} shows that sampling values according to their corresponding attention score (we call this {\em oracle sampling}) achieves a much lower (up to $4\times$) estimation error than the naive \TopK selection. 
Deploying sampling estimation in attention is promising, but three challenges remain. 
First, how a reduction of the attention error can make a difference in downstream performance is unclear~\citep{NEURIPS2019_a2ce8f17,8555143}. 
Second, modeling the attention score distribution is necessary for efficient sampling, but inferring the distribution parameters requires expensive computations. 
Third, fully leveraging the resources of modern hardware, GPU and CPU, with a theoretically efficient algorithm is non-trivial.

This paper proposes Magic samPlIng for Generation (\sys), which leverages  Locality sensitive hashing (LSH) sampling for efficient LLM generation. LSH is employed for sampling to approximate the attention score distribution and estimate attention output. By computing hash functions on GPU and conducting sampling on CPU, \sys can allow massive hash tables and hash functions compared to prior work~\citep{kitaev2020reformer,chen2021scatterbrain}, which are of vital importance for accurate estimation~\citep{8555143}. Following the practice of~\citet{aminabadi2022deepspeed,he2024fastdecode}, we offload the KV cache computation, which is memory bound, to CPU to allow a larger batch or longer context. Specifically,
\begin{itemize}
    \item In~\Cref{sec: rethinking}, we analyze the failures of \TopK attention. Moreover, we study sampling-based attention estimation assuming an oracle for the key distribution (\textbf{Oracle Sampling Estimation}) and empirically demonstrate that it is consistently more effective both for distribution estimation and downstream tasks. 
     \item In~\Cref{sec: importance sampling,sec: lsh variance,sec: approx via LSH}, we present a sampling algorithm to approximate oracle sampling for attention estimation based on locality sensitive hashing and the intuition and motivation from statistic perspectives. To our best knowledge, \sys is the first to leverage LSH sampling in self-attention in decoder-only LLM generation. 
     \item In~\Cref{sec: system codesign}, we present our system design to efficiently offload attention computation on the CPU, breaking the memory limit of the GPU for serving larger batches or longer contexts. We also overcome the new challenges of computation and memory size raised by our sampling algorithm to support a larger scale of hashing tables beyond prior work~\citep{chen2021scatterbrain,kitaev2020reformer}.
     
\end{itemize}
In~\Cref{sec:evaluations}, we show the empirical evaluation results of the performance of \sys, demonstrating the accuracy and efficiency. While maintaining high accuracy for diverse tasks, \sys can improve serving throughput by $1.5\sim5\times$ (A100, L20, RTX 4090) and can achieve 54ms decoding latency on a single RTX 4090 for Llama-3.1-8B-Instruct ~\citep{dubey2024llama} with 96K context. More importantly, we show that \sys already outperforms \TopK attention in the two aggregation tasks in \Cref{fig: topkperf}, suggesting that sampling indeed goes beyond \TopK attention.

\section{Background}
\label{sec: background}
In this section, we formulate the targeted attention estimation problem and related works. 
\subsection{Problem formulation}
\label{sec: formulation}
In LLM decoding phase, self-attention part calculates a weighted average of previous values by
\begin{equation}
    o = 
    \mathrm{Softmax}(\frac{qK^T}{\sqrt{d}})V 
    = wV
    \quad q\in \mathbb{R}^{1\times d} 
    \quad K,V \in \mathbb{R}^{n\times d}
    \quad w\in \mathbb{R}^{1\times n} 
\end{equation}
where $d$ is the head dimension and $n$ is the context size. $K = [k_1, k_2,...,k_n], V = [v_1, v_2,...,v_n], k_i, v_i \in \mathbb{R}^{1\times d}$ is KV cache. Normalized attention weight $w = \mathrm{Softmax}(\frac{qK^T}{\sqrt{d}}) \in \mathbb{R}^{1\times n}$ is also called attention (score) distribution. Our target is to find sampling matrix $\Pi \in \mathbb{R}^{n\times m}$ and diagonal matrix $D \in \mathbb{R}^{m\times m}$ which minimize  
\begin{equation}
   \delta = ||wV - w\Pi D \Pi^T V||
\end{equation}
where $m \ll n$ is computation budget. For \TopK attention, suppose $w_{r_1} > ... > w_{r_m} > ... > w_{r_n}$, then
\begin{align}
   \Pi_{i,j} = \left\{
\begin{aligned}
1 & , & \text{if } i=r_j, \\
0 & , & \text{otherwise}.
\end{aligned}
\right. \quad
D_{ii} = \frac{1}{\sum_{i=1}^{m}w_{r_i}}
\end{align}
\subsection{Related works}
\textbf{Efficient Attention.}
Attention approximation has been long studied. Reformer~\citep{kitaev2020reformer}, KDEformer~\citep{zandieh2023kdeformer} and ScatterBrain~\citep{chen2021scatterbrain} 
tackle the problem via locality sensitive hashing. These methods work in training and encoder models like BigGAN~\citep{brock2019largescalegantraining}. Theoretically, the error bounds and minimal workload required are continuously improved~\citep{brand2023algorithm,NEURIPS2023_c7286145} but have not proven to be practical for wall-clock acceleration in LLM decoding. Besides, flash-attention~\citep{flash_attn,flash_attn2,dao2022flashattention}, flash-decoding~\citep{cascade-inference,hong2024flashdecodingfasterlargelanguage} and SlimAttention~\citep{he2024inferenceperformanceoptimizationlarge} losslessly accelerate scaled product attention operator by maximizing the utilization of hardware, which is orthogonal to our approach.

\textbf{Locality sensitive hashing.}
Locality sensitive hashing (LSH)~\citep{NEURIPS2019_a2ce8f17,8555143} is a family of hashing functions which assigns the same hash codes for similar inputs with higher probability than others~\citep{chen2019slide,jafari2021survey}. LSH uses two hyper-parameters, $(K, L)$. $L$ hash tables are independently built. Each hash table has its own function $H$ which projects a high-dimension vector to an integer by concatenating $K$ random independent hash functions. In the sampling process, all vectors that share hash codes in at least one hash table with a query will be collected.
\textbf{SimHash}~\citep{10.1145/509907.509965} is the LSH family based on cosine similarity. For a vector $x \in \mathbb{R}^d$, SimHash generates a random hyperplane $w$ and returns $\mathrm{Sign}(w^Tx)$. Vectors share the same sign if and only if the random projection is not in between them. For a random projection, all angles are equally likely, thus the probability that
two vectors $x$, $y$ share the same sign is $p = 1 - \frac{\theta}{\pi}$, where $\theta = \arccos{\frac{xy^T}{||x||\cdot ||y||}}$.  If we have $L$ hash tables each with $K$ random hash functions, the probability of $y$ to be retrieved by query $x$ is $1 - (1 - p^K)^L$.

\textbf{KV Cache reduction.}
To get rid of memory bound introduced by KV cache thus enabling a larger batch size or serving a longer prompt, many methods are proposed to reduce the volume of KV cache. For example, H$_2$O~\citep{NEURIPS2023_6ceefa7b}, SnapKV~\citep{li2024snapkv} and Keyformer~\citep{adnan2024keyformer} calculate heuristics during the prefilling phase to decide which tokens to preserve for decoding phase. Quest~\citep{tang2024quest} and Loki~\citep{singhania2024loki} do not evict KV cache but apply dynamic sparsity to reduce KV Cache loading at inference time. 
Besides the reduction along the dimension of sequence length, methods like KIVI~\citep{liu2024kivi} and QServe~\citep{lin2024qserve} reduce the size of KV Cache by quantization. 
\section{Rethinking attention sparsity}
\label{sec: rethinking}
In this section,  we examine $\mathrm{TopK}$ attention, which is the theoretical upper bound of prior search-based algorithms, including both static methods~\citep{NEURIPS2023_6ceefa7b,li2024snapkv} and dynamic methods~\citep{tang2024quest,singhania2024loki,mao2024iceformer}. 
We show that \TopK is \emph{sub-optimal} and present another attention approximation based on sampling and estimation with an oracle that improves the accuracy and/or the computation cost. 

\subsection{Achilles' heel of TopK attention}
\begin{wrapfigure}{r}{150px}
    \centering
    \vspace{-10mm}
    \includegraphics[width=140px]{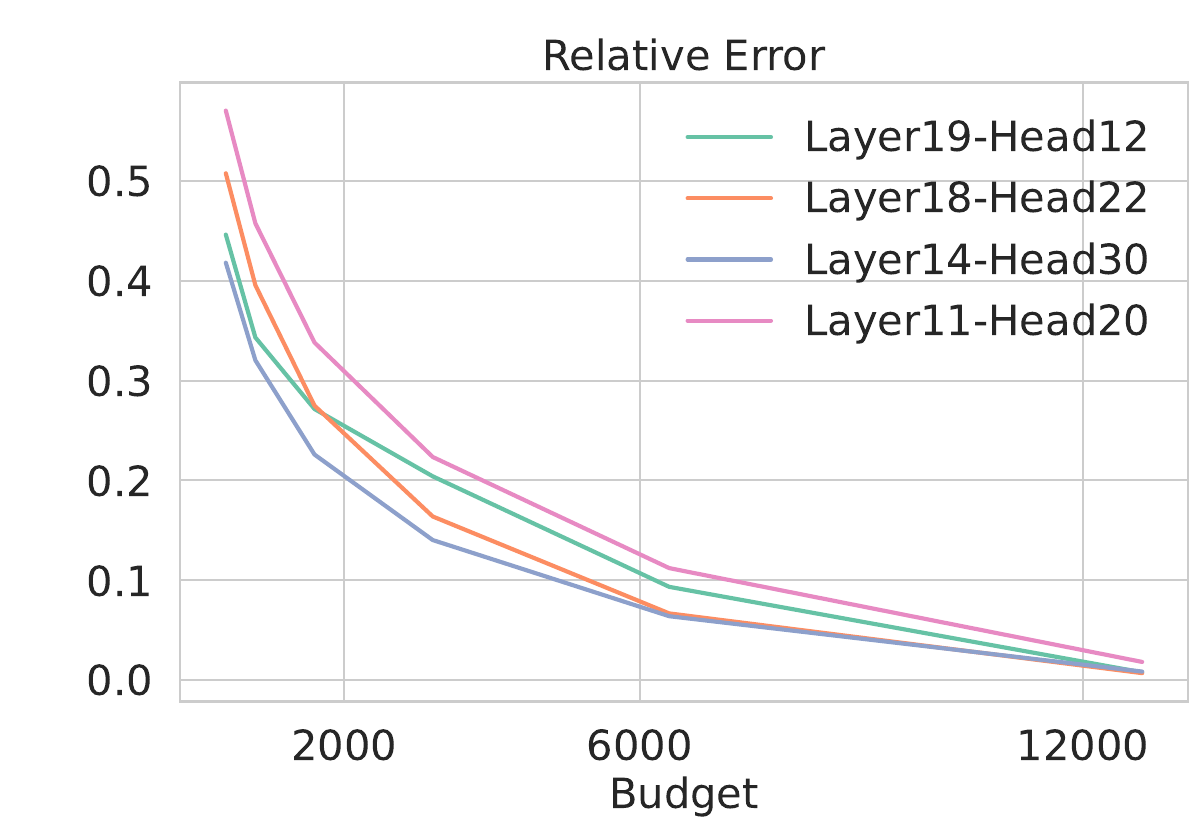}
    \caption{\TopK estimation error for a KV-cache of 16k tokens.}
    \label{fig: topkerror}
    \vspace{-5mm}
\end{wrapfigure}
As it is defined, $\mathrm{TopK}$ attention only computes the weighted average on elements with the highest attention scores. To quantify its performance, the {\em computation budget} of \TopK attention is defined as the number of selected tokens, i.e., the $\mathrm{K}$ of \TopK. 
Searching-based sparse attention algorithms, like~\citep{tang2024quest,singhania2024loki,wu2024retrieval}, are approximations for $\mathrm{TopK}$ attention by replacing the true $\mathrm{TopK}$ keys with the ones found by approximate searching algorithms. 

However,  we find significant performance degradation in downstream tasks caused by $\mathrm{TopK}$ attention as shown in~\Cref{fig: topkperf}. 
Although $\mathrm{TopK}$ attention preserves accuracy for retrieval tasks that only require a minimal subset of the context (needle-in-a-haystack single/multikey~\citep{hsieh2024ruler}), it severely degrades for aggregation tasks that leverage the full context (common word extraction and frequent word extraction~\citep{hsieh2024ruler}). 
Intuitively, the information is distributed more broadly for aggregation tasks, which results in less peak attention score distribution. 

$\mathrm{TopK}$ attention is \textit{biased} and \textit{inaccurate}, especially when the distribution of attention scores is long-tailed and the computation budget or density (i.e., $K$) is limited.  
Unfortunately, long-tailed phenomena do occur in LLMs across all layers (prior works~\citep{xiao2023efficient,tang2024quest,sun2024triforce} usually skip the first two layers to maintain accuracy) as presented in~\Cref{fig: longtail}.
$\mathrm{Top}20\%$ tokens can only cover $70\sim80\%$ attention scores, leaving a large proportion of keys and values not considered, which is translated into a non-negligible ($15\sim20\%$) estimation error in~\Cref{fig: topkerror}.

\begin{figure*}
    \centering
    \subfloat[$k_{sink}$]{
\includegraphics[width=0.31\linewidth]{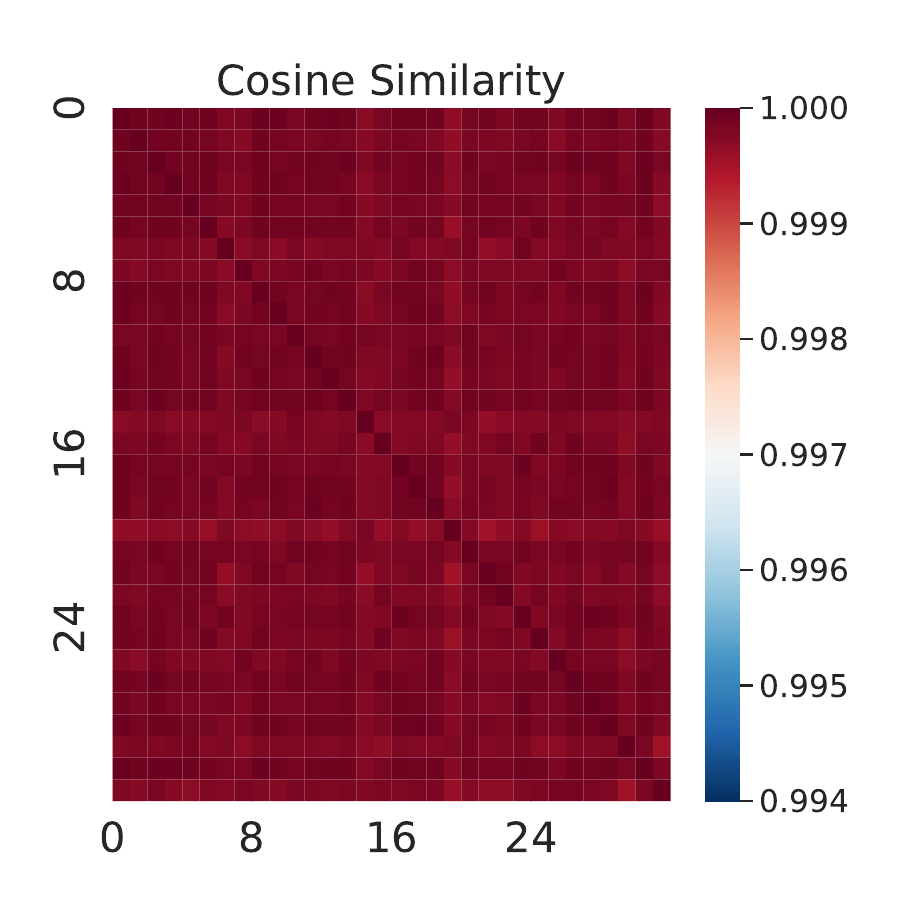}
    \label{fig: sink-k-cos}
    }
    \subfloat[$k_{avg}$]{
\includegraphics[width=0.31\linewidth]{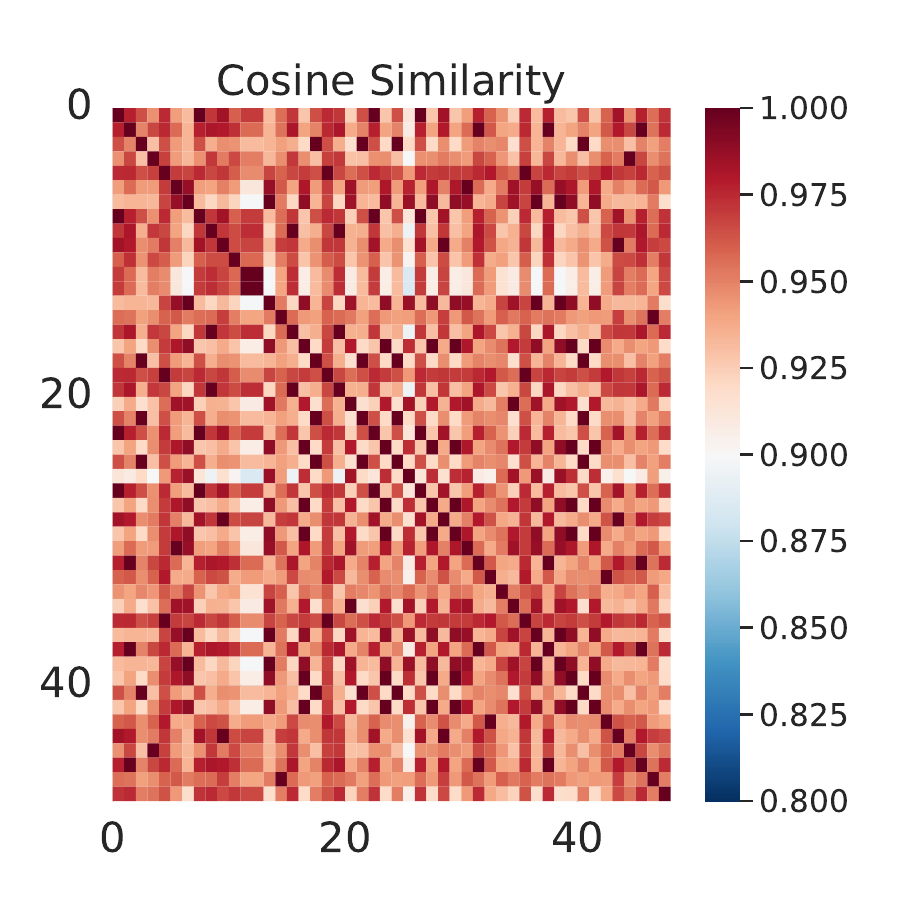}
    \label{fig: mean-k-cos}
    }
  \subfloat[Cosine similarity of $k_{sink}$ and $k_{avg}$]{
\includegraphics[width=0.31\linewidth]{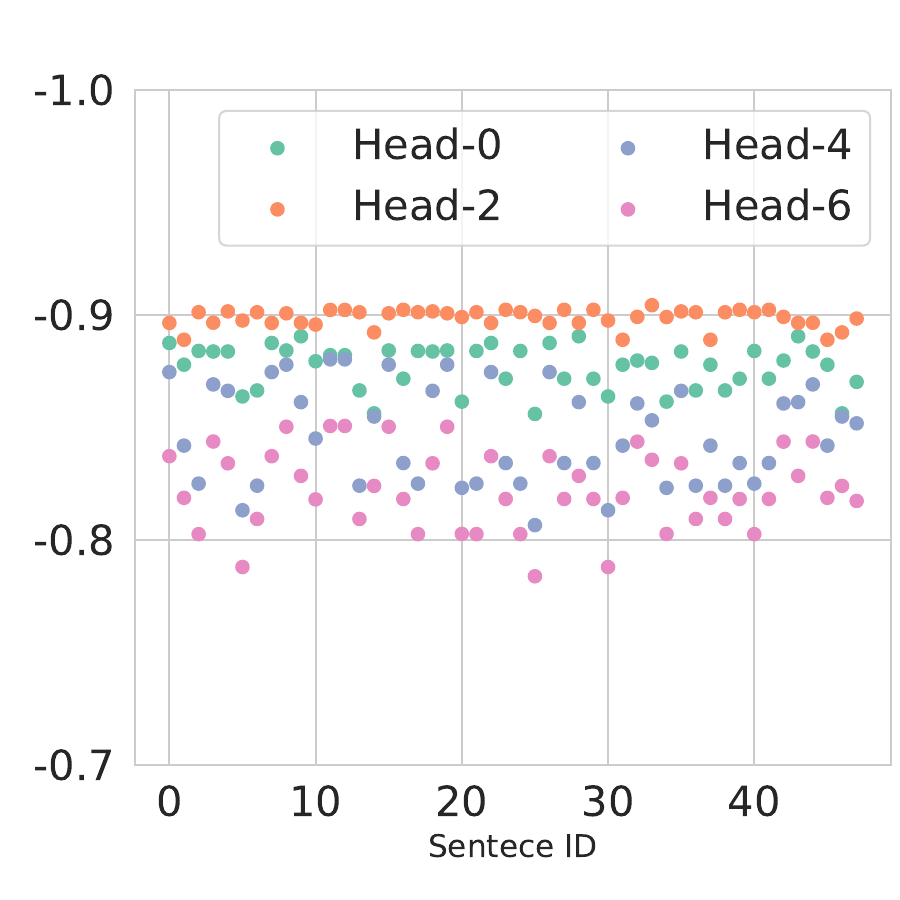}
    \label{fig: sink-mean-cos} 
    }
    \caption{Geometric information of attention. \textbf{Left:} With arbitrary input, the orientation of $k_{sink}$ almost remains the same, with a minimum similarity $>0.99$ across sampled inputs. \textbf{Mid:} The orientation of $k_{avg}$ is stable across various input sentences with a similarity $>0.9$ observed. \textbf{Right:} $k_{sink}$ and $k_{avg}$ are almost opposite with similarity between $-0.9\sim-0.8$.}
\end{figure*}

To better understand the attention distribution, we study the geometry of $q,k$ and make the following three observations. (1) Key states of the initial token (also known as attention sink, denoted by $k_{sink}$) remain almost the \textbf{same} for arbitrary input. In~\Cref{fig: sink-k-cos}, we randomly draw $32$ samples from the vocabulary and measure the mutual cosine similarity of key states. Surprisingly, we find that the orientations of the key states of different input tokens are almost \textbf{identical} with a similarity $>0.99$.
(2) The orientation of the center of key states (i.e. $k_{avg} = \frac{1}{n}\sum_{i=1}^{n}k_i$) remains \textbf{stable} for different input sentences. In~\Cref{fig: mean-k-cos}, we measure the mutual cosine similarity of $k_{avg}$ of $50$ different input sentences. Although variance exists, the similarity of $k_{avg}$ is over $0.9$. (3) The orientations of $k_{avg}$ and $k_{sink}$ are almost \textbf{opposite}. In~\Cref{fig: sink-mean-cos}, we find that for each head, $k_{sink}$ and $k_{avg}$ has a cosine similarity between $-0.9\sim-0.8$.

These observations shape the geometry as shown in~\Cref{fig: geometry}. The attention sink, which is static regardless of input, produces high sparsity in the attention distribution, whereas other parts are more uniformly distributed. Simply applying \TopK will place even more weight on the sink token, thus losing contextual information. In addition, misaligning $q$ and $k$ also causes difficulty in search \citep{liu2024retrievalattention}. 
\subsection{Estimate attention with sampling}
\label{sec: sample}
\begin{figure*}
\vspace{-8mm}
    \centering
    \subfloat[Estimation error Layer11]{
\includegraphics[width=0.32\linewidth]{figures/os_vs_topk_1.pdf}
    \label{fig: oracle sampling numerical error layer11}
    }
\subfloat[Estimation error Layer19]{
\includegraphics[width=0.32\linewidth]{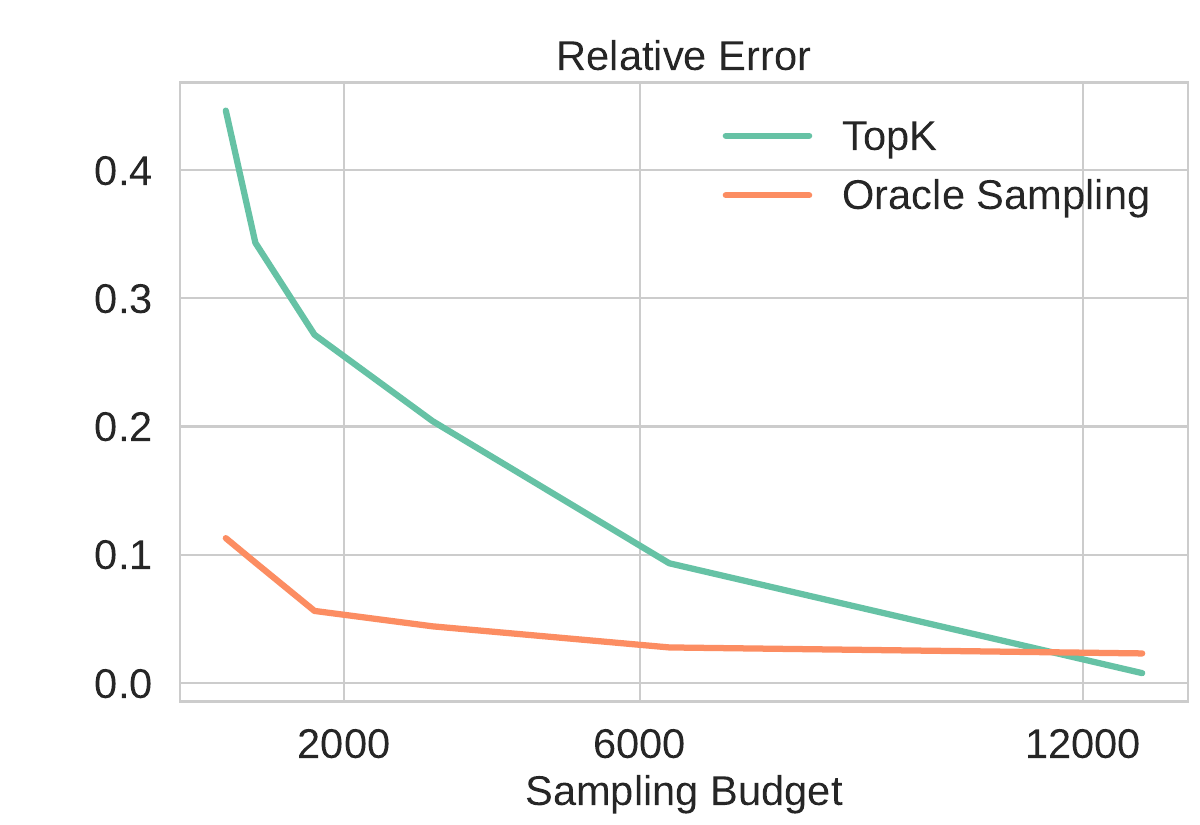}
    \label{fig: oracle sampling numerical error layer19}
    }
  \subfloat[Downstream Comparison]{
\includegraphics[width=0.3\linewidth]{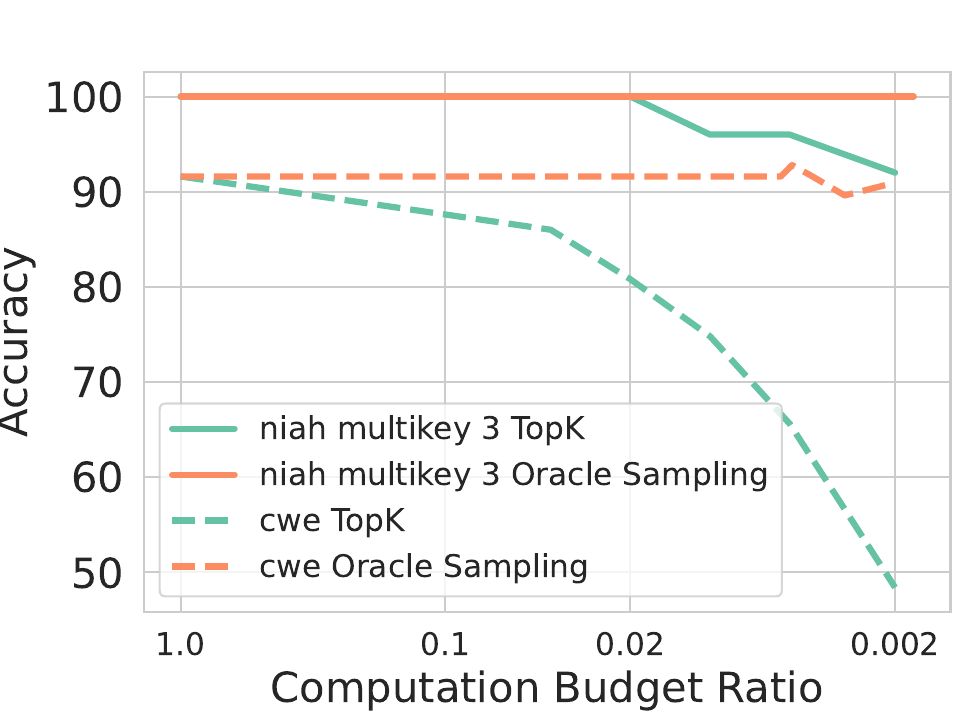}
    \label{fig: downstream comparison} 
    }
    \caption{\textbf{Left and Middle:} Oracle sampling estimation can significantly reduce numerical error compared to $\mathrm{TopK}$ attention. The evaluated context size is 16k. 
    The $x$-axis is {\em sampling budget} for oracle sampling and {\em computation budget} for $\mathrm{TopK}$ attention. Notice that the estimation error of \TopK attention will cross oracle sampling after a certain large budget (12k in figures). 
    This is because oracle sampling will repetitively sample the same subset of tokens with a high probability while \TopK will not.
    \Cref{theorem: cost} further explains this. \textbf{Right:} Downstream comparison for oracle sampling estimation and $\mathrm{TopK}$ attention. The $x$-axis for both methods is {\em computation budget ratio}, i.e. the fraction of selected/sampled tokens.}
      \vspace{-3mm}
\end{figure*}

Existing TopK attention mechanisms ignore tokens in the KV cache with low attention scores, which introduces a bias since the ignored tokens comprise a large proportion of attention scores (\Cref{fig: longtail}).
As a result, TopK attention achieves suboptimal performance for long-context tasks, such as information aggregation (\Cref{fig: topkperf}). 
Increasing the computation budget for TopK attention does help reduce the estimation error (\Cref{fig: topkerror}) since it will involve more elements in computing. However, the following question is posed:
\begin{tcolorbox}[myquote]
\itshape Can we improve the estimation quality with low computational budgets?
\end{tcolorbox}
Inspired by {\em mark and recapture}~\citep{lukacs2009closed,mcbook,lohr2021sampling,chen2018unique}, we show in the following that attention output can be estimated with sampling. 
Using notations from~\Cref{sec: formulation} we can re-write attention output $o$ as the expectation of $v_i,1\le i \le n$ from distribution $w$, i.e. $o = \mathbb{E}_{i\sim w}(v_i)$, which can be estimated by the following method.
\begin{definition}[Oracle Sampling Estimation]
Given a sampling budget $\mathcal{B}$ and normalized attention score $w$, $\mathcal{B}$ elements are sampled independently from $w$ (i.e. $i_1, i_2,...,i_{\mathcal{B}} \overset{\mathrm{iid}}{\sim}  w$). Then the attention output is estimated as
\begin{equation}
    \Bar{o} = \frac{1}{\mathcal{B}}\sum_{j=1}^{\mathcal{B}}v_{i_j}
\end{equation}
\end{definition}
This is not the lowest variance estimator but has better downstream performance (see~\Cref{sec: oracle}).
We call it ``oracle'' because it assumes that the exact attention vector $w$ is known, which is not true for sparse attention approximations. 

\begin{theorem}
\label{theorem: unbias}
Oracle sampling estimation is unbiased, and the trace of covariance monotonically decreases with $\mathcal{B}$. 
\end{theorem}

This theorem (proved in~\Cref{sec: proof}) theoretically guarantees a low estimation error of oracle sampling. 
We also present an empirical comparison between oracle sampling estimation and  $\mathrm{TopK}$ attention in~\Cref{fig: oracle sampling numerical error layer11,fig: oracle sampling numerical error layer19}. 
In summary, oracle sampling estimation can reduce relative error by up to $4\times$. 

Note that the sampling budget $\mathcal{B}$ is not the actual computation cost for oracle sampling estimation: 
duplicate $X_i$ need to be computed/loaded only once, so $\Bar{o}$ can be computed by
\begin{equation}
     \Bar{o} = \sum_{i\in \mathcal{S}} \frac{f_i}{\mathcal{B}} v_i \quad {S} = \mathrm{Unique}(\{i_{1\le i \le \mathcal{B}}\})
\end{equation}
where $f_i$ is the number of duplicates of $X_i$. 
Intuitively, if $w$ has an peaked distribution (e.g. $w_i >99\%$), then almost all samples in $\{i_1,...,i_\mathcal{B}\}$ are identical to $i$. 
The actual computation cost of oracle sampling estimation is $|{S}|$, the number of \emph{unique} samples, which we bound in the following:
\begin{theorem}
\label{theorem: cost}
The expected computation budget ($\mathbb{E}({|{S}|}))$ has an upper bound of $1 + \mathcal{B}\epsilon$, where $\epsilon = 1 - \max_{i}w_i$.
\end{theorem}

This theorem (proved in~\Cref{sec: proof}) shows that the computation cost of oracle sampling is usually far less than the sampling budget. 
In~\Cref{fig: downstream comparison}, we present the downstream accuracy comparison between oracle sampling estimation and $\mathrm{TopK}$ attention. 
The former preserves high accuracy for both tasks, even with a very small computation cost ($0.002\%$ out of 16k context, which is approximately $32$). In~\Cref{sec:sampling example}, we provide an intuitive example to explain why sampling outperforms \TopK in estimation.

\section{\sys}
\label{sec:magic}

\cref{sec: sample} demonstrates the potential of sampling-based estimation. In \Cref{sec: importance sampling,sec: lsh variance}, we present how we arrive at Locality sensitive hashing to unleash this potential from a statistical perspective. 
In \Cref{sec: approx via LSH}, we show the practical algorithm. 
Finally, in \Cref{sec: system codesign}, we demonstrate our system co-design for accurate and efficient LLM decoding through GPU-CPU collaboration. 

Note that most of the derivations in this section might be classical and can even be found in textbooks, but our goal is to leverage them to motivate \sys design and precisely demonstrate the power of a rigorously sound algorithm with system co-design in deep generative models.
\subsection{Self-normalized importance sampling for attention estimation}
\label{sec: importance sampling}
Oracle sampling estimation cannot go beyond $2\times$ wall clock speed up because obtaining distribution $w$ requires full computation of all $qk_i^T$, thereby only saving the $wV$ computation.

Fortunately, importance sampling~\citep{kloek1978bayesian,mcbook,lohr2021sampling} allows us to estimate unknown distribution $w$ by sampling from a proposed distribution $u$. 
In our problem setting, the normalization factor of $w$, i.e. $Z = \sum_{i=1}^{n}\exp{\frac{qk_i^T}{\sqrt{d}}}$ is also unknown because computing it requires evaluating all $qk_i^T$. 
However, we do have access to unnormalized weights $\widetilde{w_i} = e^{\frac{qk_i^T}{\sqrt{d}}}$ for sampled indices $i$. 
\newcommand{\budget}{\mathcal{B}}
Hence, by employing a variant of importance sampling, \textbf{self-normalized importance sampling}~\citep{mcbook}, we sample indices $i_1, i_2,...,i_\budget$ from a proposed distribution $u$ and the resulting estimator is
\begin{equation}
    X^\mathrm{IS} = 
    \frac{1}{\widetilde{Z}}\sum_{j=1}^{\budget} \frac{\widetilde{w_{i_j}}}{u_{i_j}}v_{i_j}
\quad
\text{where}
\quad 
\widetilde{Z} = 
    \sum_{j=1}^{\budget}\frac{\widetilde{w_{i_j}}}{u_{i_j}}
    \label{eq: self normalized is}
\end{equation}
which has a very nice property for accurately estimating attention output that $\mathbb{P}[\lim_{\budget \rightarrow \infty} X^\mathrm{IS} = o] = 1$. 


Its variance\footnote{We assume head dimension $d=1$ here for simplicity. Higher dimensions have similar formulations and analyses by replacing variance with the trace of covariance. } is related to the distribution $u$, and can be approximated by
\begin{equation}
\label{eq: variance}
   \widetilde{\mathrm{Var}}(X^\mathrm{IS}) = 
   \frac{1}{\budget}\mathbb{E}_{i\sim u}[\frac{w_i^2}{u_i^2}(v_i - o)^2] = 
   \frac{1}{\budget Z^2}\mathbb{E}_{i\sim u}[\frac{\widetilde{w_i}^2}{u_i^2}(v_i - o)^2]
\end{equation}
To minimize the variance, $u$ should satisfy $u \propto \widetilde{w_i}|v_i - o|$~\citep{article}.
The variance will be high if $u_i$ and $\widetilde{w_i}|v_i - o|$ assign a high probability mass to different regions of the sample space or have different modes. 
Therefore, the challenge is computing a distribution $u$ aligned with $\widetilde{w_i}|v_i - o|$ without accessing too many $\widetilde{w_i}$.  Besides, \Cref{eq: self normalized is} requires that sampling probability $u$ can be computed and $u_i > 0$, which is not satisfied by many deterministic approximations like \TopK.

\subsection{Variance reduction with LSH}
\label{sec: lsh variance}

We decompose $\widetilde{w_i}|v_i - o| = \exp({\frac{qk_i^T}{\sqrt{d}} + \log|v_i - o|}$). 
We observe empirically (\Cref{fig:fluctuate} in the appendix) that $\log|v_i - o|$ does not fluctuate significantly compared to $\frac{qk_i^T}{\sqrt{d}}$. Hence, we simplify the requirement of $u$ to share the same peaks with $qk_i^T$. By the following transformation,
\begin{align}
    r = \max_{1\le i \le n}|k_i|  \quad 
    \bar{q} = [q, 0] \quad 
    \bar{k_i} = [k_i, \sqrt{r^2 - |k_i|^2}]
    \label{eq: data transform}
\end{align}
we further transfer the inner product $qk_i^T$ to cosine similarity between $\bar{q}$ and $\bar{k_i}$ (which is a common practice in Maximum Inner Product Search~\citep{shrivastava2014asymmetric}).


Inspired by prior work~\citep{spring2017new,MLSYS2020_ca3480d8}, we leverage Locality sensitive hashing-based sampling for this estimation problem. 
Specifically, leveraging a hash function $h$ in the LSH family that preserves cosine similarity such as SimHash~\citep{Sadowski2007SimHashH}, we can sample from probability distribution $u_i = \mathbb{P}[{h}(q) = h(k_i)]$ which is monotonic to $\cos{\frac{qk_i^T}{|q|\cdot |k_i|}}$. 


\subsection{Algorithm Design}

\label{sec: approx via LSH}
To make this estimation practical, \sys is implemented by the following specific design.

\textbf{Estimator approximation.} Self-normalized important sampling \Cref{eq: self normalized is} requires $i_1,i_2,...,i_k$ $\mathrm{iid}$ sampled, but the probabilities provided by hashing are not normalized.
Hence we adapt our estimator: After obtaining $S$ with probability $u$, \sys computes 
    \begin{equation}
    X = \frac{\sum_{i=1}^{n} \frac{\widetilde{w_i}}{u_i}v_i\mathbf{1}_{i\in S}}{\sum_{i=1}^{n}\frac{\widetilde{w_i}}{u_i}\mathbf{1}_{i\in S}} = \frac{\sum_{i\in S} \frac{\widetilde{w_i}}{u_i}v_i}{\sum_{i\in S}\frac{\widetilde{w_i}}{u_i}}
    \label{eq: importance sampling unormalized}
\end{equation}

\textbf{Hash function selection.} \sys leverages \textbf{SimHash}~\citep{Sadowski2007SimHashH}, that draws with $K \times L$ random vectors. 
For each of the $L$ hash tables, the ${q}$ and ${k_i}$s vectors are projected on $K$ directions, and only the sign of the projection is kept, which yields a $K$-bit hash value. Key ${k_i}$ is sampled only if there exist at least \textbf{two}\footnote{Empirical results show that requiring hits in two hash tables greatly improves accuracy over standard SimHash.} hash tables where ${k_i}$ shares the hash value with ${q}$. 
The corresponding probability is
\begin{equation}
u_i = \mathbb{P}[\text{${k_i}$ is sampled}] = 1 - (1 - p^{K})^{L} - Lp^K (1 - p^K)^{L-1} 
\quad\text{where}\quad
p = 1 - \frac{1}{\pi}\arccos{\frac{{q}{k_i}^T}{|{q}|\cdot |{k_i}|}}
\label{eq: lsh sampling probability}
\end{equation}

\begin{wrapfigure}{}{0.5\textwidth}
  \vspace{-0.1in}
    \begin{minipage}{0.5\textwidth}
\begin{algorithm}[H]
\caption{\Sys Decoding}
\label{alg:magicpig}
\begin{algorithmic}
\State \textbf{Input:} $\boldsymbol{K}, \boldsymbol{V} \in {R}^{n\times d}$, $\boldsymbol{q} \in {R}^{1\times d}$, random projectors $\boldsymbol{W}\in R^{d\times (K \times L)}$, hash tables $\boldsymbol{HT}$, static KV cache $\boldsymbol{K_T}, \boldsymbol{V_T} \in {R}^{t\times d}$.
\State {\color{commentcolor}{\textit{Compute hash code for new query}}}
\State $\boldsymbol{q}_{\text{code}} = \textbf{Encode}(\boldsymbol{q},\boldsymbol{W})$ 

\State {\color{commentcolor}{\textit{Query hash tables to sample $\boldsymbol{S}$ in~\Cref{eq: importance sampling unormalized}}}}
\State {$\boldsymbol{S}$ = $\textbf{Query}(\boldsymbol{HT}, \boldsymbol{q}_{\text{code}}), \boldsymbol{K_S} = \boldsymbol{K[S]},  \boldsymbol{V_S} = \boldsymbol{V[S]}$}
\State {\color{commentcolor}{\textit{Compute inner product for $\boldsymbol{q}$ and sampled $\boldsymbol{K}$}}}
\State{$\boldsymbol{w_S} = \boldsymbol{q}\boldsymbol{K^T_S}$, $\boldsymbol{w_T} = \boldsymbol{q}\boldsymbol{K^T_T}$}
\State {\color{commentcolor}{\textit{Compute collision probability for each hash function}}}
\State{$\boldsymbol{p} = 1 - \frac{1}{\pi}\arccos({{\boldsymbol{w}} / {(||\boldsymbol{q}|| \cdot ||\boldsymbol{K_S}||})})$} 
\State {\color{commentcolor}{\textit{Compute sampling probability}}}
\State{$\boldsymbol{u} = 1 - (1 - \boldsymbol{p}^{K})^{L} - L\boldsymbol{p}^K (1 - \boldsymbol{p}^K)^{L-1}$} 
\State {\color{commentcolor}{\textit{Compute attention output estimation}}}
\State{$\boldsymbol{\Bar{o}} = \textbf{Softmax}(\frac{[\boldsymbol{w_S}, \boldsymbol{w_T}]}{\sqrt{d}} - \log([\boldsymbol{u}, \boldsymbol{1_{t}}]))[\boldsymbol{V_S}, \boldsymbol{V_T}]$}
\State {\textbf{Return} $\boldsymbol{\Bar{o}}$} 
\end{algorithmic}
\end{algorithm}
    \end{minipage}
    \vspace{-0.8in}
\end{wrapfigure}

\textbf{Data pre-processing.} Before building hash tables, \sys centers the $k_i$ vectors. 
As shown in~\Cref{fig: geometry}, keys are almost always concentrated on one side of the queries, except the initial token. 
In this case, random projections cannot effectively distinguish keys, resulting in uniform sampled probabilities. $\mathrm{Softmax}$ is translation invariant. 
Centering ($\bar{k_i} = k_i - \frac{1}{n}\sum_{i=1}^{n}k_i$) distributed the keys better and remains computationally equivalent. 
Combining~\Cref{eq: importance sampling unormalized,eq: lsh sampling probability} gives a closed form of the \sys attention estimation. Assuming sample set $S$ is obtained with LSH, 
\begin{align}
    \Bar{o} &= \sum_{i\in S}\frac{\exp{(\frac{qk_i^T}{\sqrt{d}} -\log u_i)}}{\sum_{i\in S}\exp{(\frac{qk_i^T}{\sqrt{d}} -\log u_i)}}  v_i \nonumber \\
    u_i & = 1 - (1 - p_i^{K})^{L} - Lp_i^K (1 - p_i^K)^{L-1}\nonumber \\
    p_i &= 1 - \frac{1}{\pi}\arccos{\frac{{q}{k_i}^T}{|{q}|\cdot |{k_i}|}} 
    \label{eq: close form}
\end{align}

\subsection{System co-design}
\label{sec: system codesign}
\begin{figure*}
\vspace{-8mm}
    \centering
    \subfloat[Memory Hierarchy]{
\includegraphics[width=0.45\linewidth]{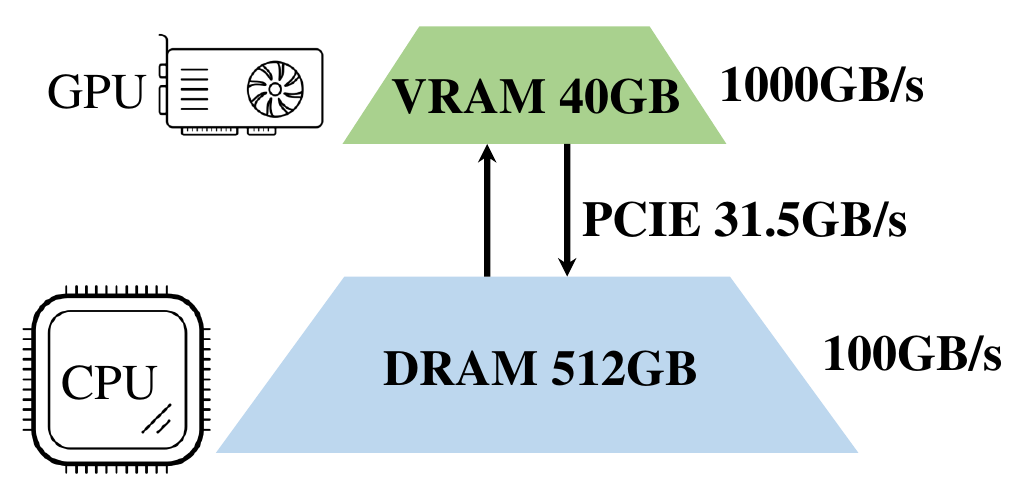}
    \label{fig: bandwidth compare}
    }
  \subfloat[Workload Partition]{
\includegraphics[width=0.45\linewidth]{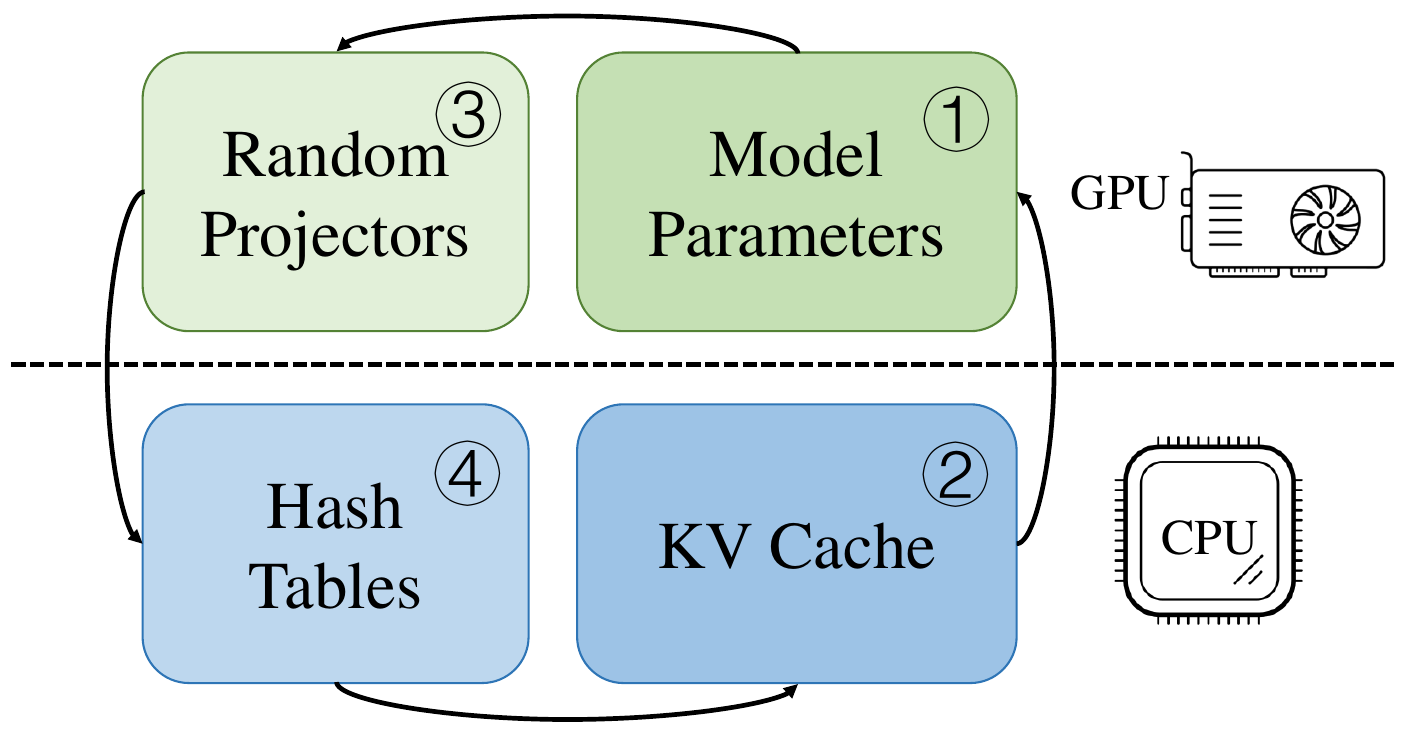}
    \label{fig: memory layout} 
    }
    \caption{\textbf{Left:} Memory hierarchy of hardware. GPU VRAM has high bandwidth but is limited. CPU DRAM is sufficient but is relatively slow. The limited bandwidth of PCIE forbids large-scale data transfer. \textbf{Right:} Workload partition of \sys. Linear projections and hash function computation (by random projection) are done on GPU, while sampling with hash tables and attention are done on CPU. The execution order is $\textcircled{1}\textcircled{3}\textcircled{4}\textcircled{2}$ at decoding time.}
    \vspace{-3mm}
\end{figure*}
The memory size of KV cache remains a bottleneck for long-context LLM decoding, especially when GPU VRAM is limited. DRAM on the CPU side offers sufficient memory capacity with $100-200$GB/s bandwidth, which is usually $10-20\%$ of GPU VRAM bandwidth  (see \Cref{fig: bandwidth compare}). 
Ideally, this gap can be mitigated by $5-10\times$ sparsity. 
To make CPU DRAM  an \textit{aggregated memory} for GPU, the workload must be partitioned. 
In our experiments, $K=9$ or 10, and $L$ is a few hundred.

Our system design extends prior work~\citep{he2024fastdecode,aminabadi2022deepspeed} by splitting LLM decoding into three parts. 
(1) Parameter computations, i.e., all linear projectors including MLP and $W_Q, W_K, W_V, and W_O$ in the self-attention module run on GPU. 
(2) Attention computation, which involves $o = \mathrm{Softmax}(\frac{qK^T}{\sqrt{d}})V$, runs on CPU. 
(3) Random projections. At generation time, for each $q$, $K \times L$ random projections are conducted to obtain the hash codes. 
Since all heads can share the same random projectors, the memory overhead is limited (400~KB in our implementation), so this step is compute-bound. 
Therefore, the projection is placed on GPU. 
(4) Retrieval. The hash codes of $q$, need to be looked up in $L$ hash tables, which is negligible computationally. However, the pre-built hash tables for $k_i$s can occupy considerable memory, making it a better fit for the CPU.
With the above partition, we are able to support hash tables with $K$ and $L$ beyond the scale of prior work~\citep{kitaev2020reformer,chen2021scatterbrain,zandieh2023kdeformer} without worrying about computation for hash codes as well as the storage of hash tables. 

\textbf{On-device cache.} Sink tokens (the first several tokens) and local tokens are more likely to be sampled according to their high similarity to the query. To further reduce CPU workload, \sys stores these tokens on GPU and does not apply LSH sampling to them. We leverage the recursive attention technique \citep{cascade-inference} to merge the attention output from CPU and GPU.

Our algorithm applies to a single attention head, see~\Cref{alg:magicpig}. 
The details of $\textbf{Encode}$, $\textbf{Query}$, as well as the hash table construction, are described in prior work~\citep{Sadowski2007SimHashH,chen2019slide}. In~\Cref{sec: hyper-parameter selection}, we discuss the selection of LSH hyper-parameter (K, L). 
\section{Evaluation}
\label{sec:evaluations}
In this section, we aim to demonstrate that \sys can speed up LLM decoding while preserving high accuracy. We first present \sys's accuracy in downstream tasks, followed by our end-to-end system results showing wall-clock performance. 
\begin{itemize}[itemsep=0.0pt,topsep=0pt,leftmargin=*]
    \item In~\Cref{sec: accuracy}, we demonstrate \sys preserves high accuracy (less than $2\%$ degradation) across moderate to long context tasks with computation cost $2\%\sim5\%$ of full attention. 
    \item In~\Cref{sec: efficiency}, we demonstrate the system performance of \sys, which achieves up to $5\times$ throughput improvement and 54ms decoding latency on a single RTX 4090 for Llama-3.1-8B-Instruct with 96K context.
    \item In~\Cref{sec: ablation}, we verify the effectiveness of centering, which is of vital importance for the success of sampling. Also, we demonstrate that \sys already outperforms \TopK attention in the two aggregation tasks in \Cref{fig: topkperf}, indicating that sampling indeed goes beyond \TopK attention.
\end{itemize}
\subsection{\sys Preserves Accuracy}
\label{sec: accuracy}
We demonstrate that \sys can preserve accuracy in diverse tasks with less than $5\%$ computation.

\textbf{Setup.} Our experiments are based on Llama~\citep{llama3modelcard,dubey2024llama,touvron2023llama} models. Three types of tasks are included, which are 3 mid-context comprehensive tasks from lm-eval-harness~\citep{eval-harness} (GSM8K-CoT~\citep{cobbe2021training}, MMLU-Flan-Cot-Fewshot~\citep{hendrycks2020measuring} and COQA~\citep{reddy2019coqa}), and 6 long context tasks from~\citep{bai2023longbench} (QASPER~\citep{dasigi2021dataset}, LCC, Repobench-P~\citep{liu2023repobenchbenchmarkingrepositorylevelcode}, TriviaQA~\citep{joshi2017triviaqalargescaledistantly}, PRE and TREC~\citep{li-roth-2002-learning,hovy-etal-2001-toward}) and 13 synthetic tasks from RULER~\citep{hsieh2024ruler} (with 50 examples per task). 

\textbf{Baselines.} Besides full attention, Quest~\citep{tang2024quest} and its variants are used as baselines. 
In its default setting, Quest uses a ``page size'' of 16, i.e. 1/16 of the full attention cost. 
To compare the methods fairly in the low computation budget regime, we also evaluate Quest with page size 32 and 64 and make sure at least one page is selected in every test example. 
The initial 4 tokens and local 64 (for LongBench~\citep{bai2023longbench} and RULER~\citep{hsieh2024ruler}) or 24 (for lm-eval-harness~\citep{eval-harness}) tokens as well as layer-$\{0,16\}$ are statically preserved. 
We do not use the theoretical transformations in~\Cref{eq: data transform} in our implementations, as we do not find them to contribute to accuracy improvements.

\textbf{Cost.} The cost for the attention approximation consists of two parts: 
$\text{Cost}_1$ is the sampling/search cost to obtain $S$ in~\Cref{eq: close form}, 
$\text{Cost}_2$ is the attention computation cost, see~\Cref{eq: close form}.  
We report the ratio of the number of FLOPs compared to the full attention computation.
For \sys, $\text{Cost}_1 \simeq 0$ and $\text{Cost}_2$ is empirically measured for different LSH hyper-parameters. For Quest with page size $K$, $\text{Cost}_1 = \frac{1}{K}$ and $\text{Cost}_2$ is  controlled manually. 


\textbf{Analysis.} From~\Cref{tab:lm-eval-harness,tab:longbench,tab:ruler}, 
(1) \sys preserves high accuracy (degradation less than $2\%$) for all kinds of tasks, with a computation cost of $2\%\sim5\%$. 
(2) Compared with Quest, which also shows reasonable performance on long context tasks, \sys also demonstrates good performance on tasks with moderate context sizes in lm-eval-harness~\citep{eval-harness}, indicating a more robust performance in general serving. (3) With LSH sampling, which introduces an order of magnitude lower sampling/searching cost ($\text{Cost}_1 $), \sys can achieve equivalent or better accuracy with only half of the computation cost.

\begin{table}[t]
\vspace{-5mm}
\centering
\caption{Comprehensive tasks on lm-eval-harness \citep{eval-harness}. \Sys significantly outperforms other methods with lower computation. The config (K, L) is a hyper-parameter of LSH for \Sys or page size and ratio of selected pages for Quest~\citep{tang2024quest}. $\text{Cost}_1$, $\text{Cost}_2$ represents the cost for searching/sampling and sparse attention computation.}
  \vspace{-2mm}
\setlength{\tabcolsep}{3.5pt}
\small
\begin{tabular}{lc|rrr|r|rr|r}
\toprule
Methods & Config & GSM & COQA & MMLU & Avg. &$\text{Cost}_1$&$\text{Cost}_2$&$\text{Cost}_{total}$.\\ \midrule
\textit{Llama-2-7b-chat} & Full & 22.4&75.8 & 49.2& 49.1 & 0.00& 1.00& 1.00\\
\rowcolor{pink!20} \Sys & (10,220) & 17.3&76.4 & 48.6& \textbf{47.4} & 0.00& 0.04& 0.04\\
\rowcolor{pink!20} \Sys & (8,90) & 18.7&75.0 & 47.9& 47.2 & 0.00& 0.08& 0.08\\
Quest & (16,0.05) & 13.0&69.4 & 41.4& 41.3 & 0.06& 0.05& 0.11\\
Quest & (32,0.1) & 15.7&70.2 & 44.0& 43.3 & 0.03& 0.10& 0.13\\
\midrule
\textit{Llama-3.1-8B-Instruct} & Full & 77.6&78.5 & 65.2& 73.7 & 0.00& 1.00& 1.00\\
\rowcolor{pink!20} \Sys & (10,220) & 72.7&78.1 & 62.7& \textbf{71.2} & 0.00& 0.03& 0.03\\
\rowcolor{pink!20} \Sys & (8,90) & 71.0&78.0 & 61.3& 70.1 & 0.00& 0.07& 0.07\\
Quest & (16,0.05) & 57.9&64.6 & 42.5& 55.0 & 0.06& 0.05& 0.11\\
Quest & (32,0.1) & 64.5&65.0 & 48.0& 59.2 & 0.03& 0.10& 0.13\\
  \bottomrule
\end{tabular}
\label{tab:lm-eval-harness}
\end{table}

\begin{table}
\centering
\caption{Long context tasks on LongBench~\citep{bai2023longbench}. \Sys preserves high accuracy with low computation. Config and cost are defined as in~\Cref{tab:lm-eval-harness}. Code models are only evaluated by Repobench-P and LCC.}
  \vspace{-2mm}
\setlength{\tabcolsep}{3.5pt}
\small
\begin{tabular}{lc|rrrrrr|r|rr|r}
\toprule
Methods & Config & QaS & RbP & LCC & PrE& TrC & TrQ & Avg. &$\text{Cost}_1$&$\text{Cost}_2$&$\text{Cost}_{total}$.\\ \midrule
\textit{Llama-3.1-8B-Instruct} & Full & 44.9&52.1 & 66.8& 100.0 &71.3&91.8& 71.2& 0.00& 1.00& 1.00\\
\rowcolor{pink!20} \Sys & (10,150) & 43.2&50.2 & 64.4&100.0&71.3&92.2& 70.3 & 0.00& 0.02& 0.02\\
\rowcolor{pink!20} \Sys & (8,75) & 43.5&50.4 & 67.0&100.0&71.7&91.7& \textbf{70.7} & 0.00& 0.05& 0.05\\
Quest & (16,0.05) & 45.7&49.7& 64.9&100.0&71.7&91.5& 70.6 & 0.06& 0.05& 0.11\\
Quest & (32,0.1) & 44.4&50.5 & 65.1&100.0&71.3&91.6& 70.5 & 0.03& 0.10& 0.13\\ \midrule
\textit{Code-Llama-13b-16K} & Full & &58.5 & 74.7& &&& 66.6& 0.00& 1.00& 1.00\\
\rowcolor{pink!20} \Sys & (10,150) & &56.9 & 74.0&&&& \textbf{65.5} & 0.00& 0.03& 0.03\\
Quest & (16,0.05) & &56.4 & 74.4&&&& 65.4 & 0.06& 0.05& 0.11\\
  \bottomrule
\end{tabular}
\label{tab:longbench}
\end{table}
\begin{table}[htbp]
\centering
\caption{Synthesized tasks on RULER~\citep{hsieh2024ruler}. \Sys preserves high accuracy with low computation. Config and cost are defined as in~\Cref{tab:lm-eval-harness}.}
\setlength{\tabcolsep}{3.5pt}
\small
\begin{tabular}{lc|rrrr|r|rr|r}
\toprule
Methods & Config & 16K & 32K & 64K & 96K & Avg. &$\text{Cost}_1$&$\text{Cost}_2$&$\text{Cost}_{total}$.\\ \midrule
\textit{Llama-3.1-8B-Instruct} & Full & 94.2&91.5 & 86.1& 83.0 & 88.7& 0.00& 1.00& 1.00\\
\rowcolor{pink!20} \Sys & (10,150) & 91.8&88.9 & 84.8&80.0& 86.4 & 0.00& 0.02& 0.02\\
\rowcolor{pink!20} \Sys & (9,120) & 93.4&90.6 & 84.7&81.5& \textbf{87.6} & 0.00& 0.04& 0.04\\
\rowcolor{pink!20} \Sys & (8,75) & 92.9&90.2 & 84.9&81.7& 87.4 & 0.00& 0.05& 0.05\\
Quest & (16,0.04) & 86.3&85.4& 81.9&74.9& 82.1 & 0.06& 0.04& 0.10\\
Quest & (32,0.06) & 84.3&84.0 & 80.1&74.4& 80.7 & 0.03& 0.06& 0.09\\
Quest & (64,0.08) & 85.2&84.3 & 77.0&74.2& 80.2 & 0.02& 0.08& 0.10\\
  \midrule
\textit{MegaBeam-Mistral-7B-512K} & Full & 91.7&88.1 & 83.5& 83.7 & 86.8& 0.00& 1.00& 1.00\\
\rowcolor{pink!20} \Sys & (10,150) & 89.8&86.5 & 81.7&80.7& 84.7 & 0.00& 0.02& 0.02\\
\rowcolor{pink!20} \Sys & (9,120) & 90.7&88.5 & 82.9&82.4& \textbf{86.1} & 0.00& 0.04& 0.04\\
\rowcolor{pink!20} \Sys & (8,75) & 90.6&86.4 & 82.8&81.6& 85.4 & 0.00& 0.05& 0.05\\
Quest & (16,0.04) & 83.3&83.2& 79.3&78.6& 81.1 & 0.06& 0.04& 0.10\\
Quest & (32,0.06) & 81.5&80.8 & 76.7&74.4& 78.4 & 0.03& 0.06& 0.09\\
Quest & (64,0.08) & 79.6&77.5 & 73.8&73.7& 76.1 & 0.02& 0.08& 0.10\\
\midrule
\textit{Llama3-8B-Prolong-512K} & Full & 93.5&90.8 & 85.1& 83.5 & 88.2& 0.00& 1.00& 1.00\\
\rowcolor{pink!20} \Sys & (10,150) & 88.0&86.4 & 81.3&78.8  & 83.6 & 0.00& 0.02& 0.02\\
\rowcolor{pink!20} \Sys & (10,170) & 89.0&88.7 & 82.8&80.0  & 85.1 & 0.00& 0.025& 0.025\\
\rowcolor{pink!20} \Sys & (9,120) & 91.4&88.2 & 82.4&80.4 & 85.6 & 0.00& 0.04& 0.04\\
\rowcolor{pink!20} \Sys & (8,75) & 91.4&88.6 & 83.1&80.5 & \textbf{85.9} & 0.00& 0.05& 0.05\\
Quest & (16,0.04) & 84.9&83.7& 78.7&78.6 & 81.5 & 0.06& 0.04& 0.10 \\
\bottomrule
\end{tabular}
\label{tab:ruler}
  \vspace{-2mm}
\end{table}
\vspace{-3mm}
\subsection{\sys Shows Impressive Efficiency across Various Hardware Settings}
We show \sys can bring up to $5\times$ wall clock speed up and reduce GPU memory consumption on different models and hardware settings (A100, L20, RTX4090). 
\begin{figure*}
\vspace{-3mm}
    \centering
    \subfloat[A100 with 34B model]{
\includegraphics[width=0.32\linewidth]{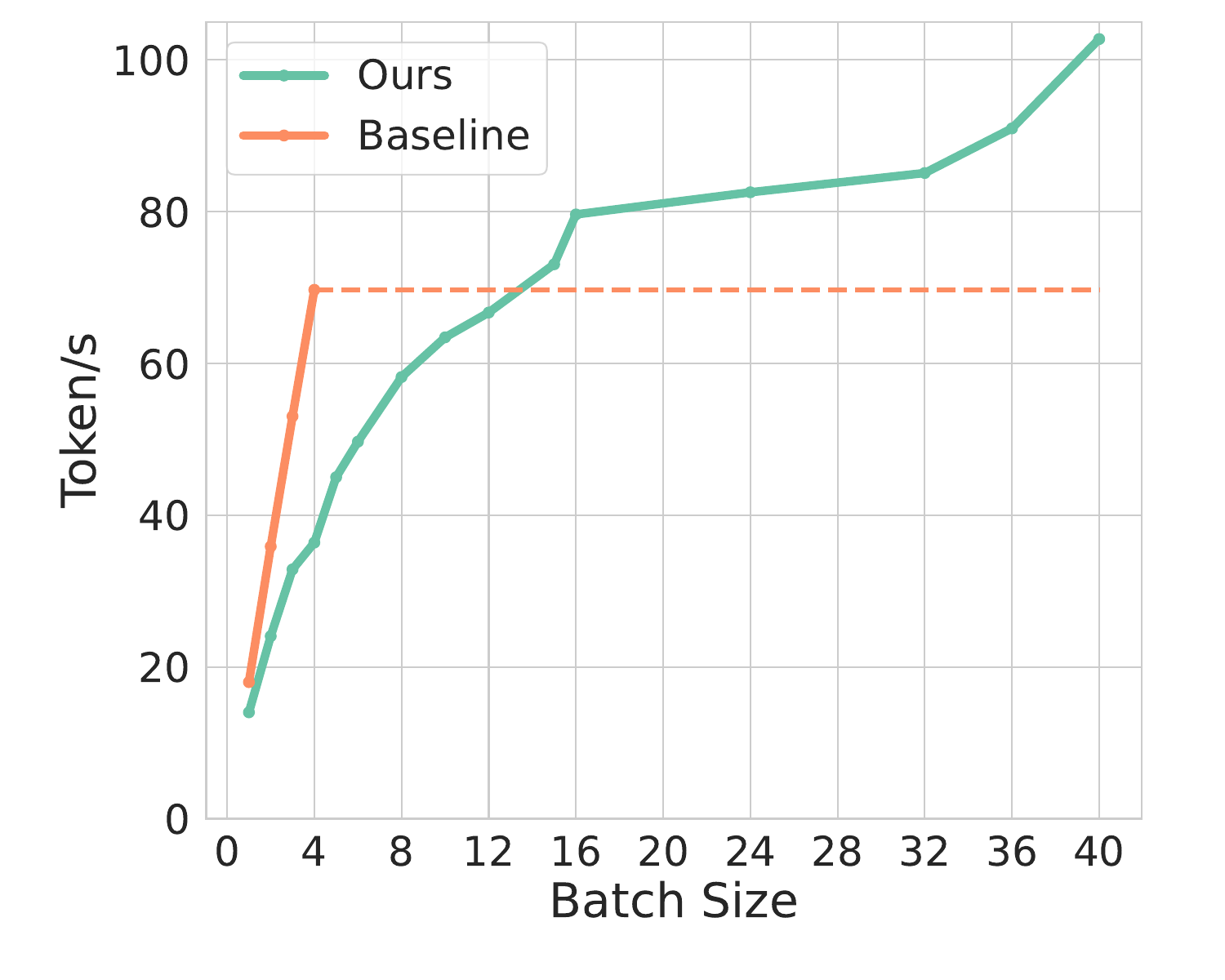}
    \label{fig: A100}
    }
\subfloat[L20 with 13B model]{
\includegraphics[width=0.32\linewidth]{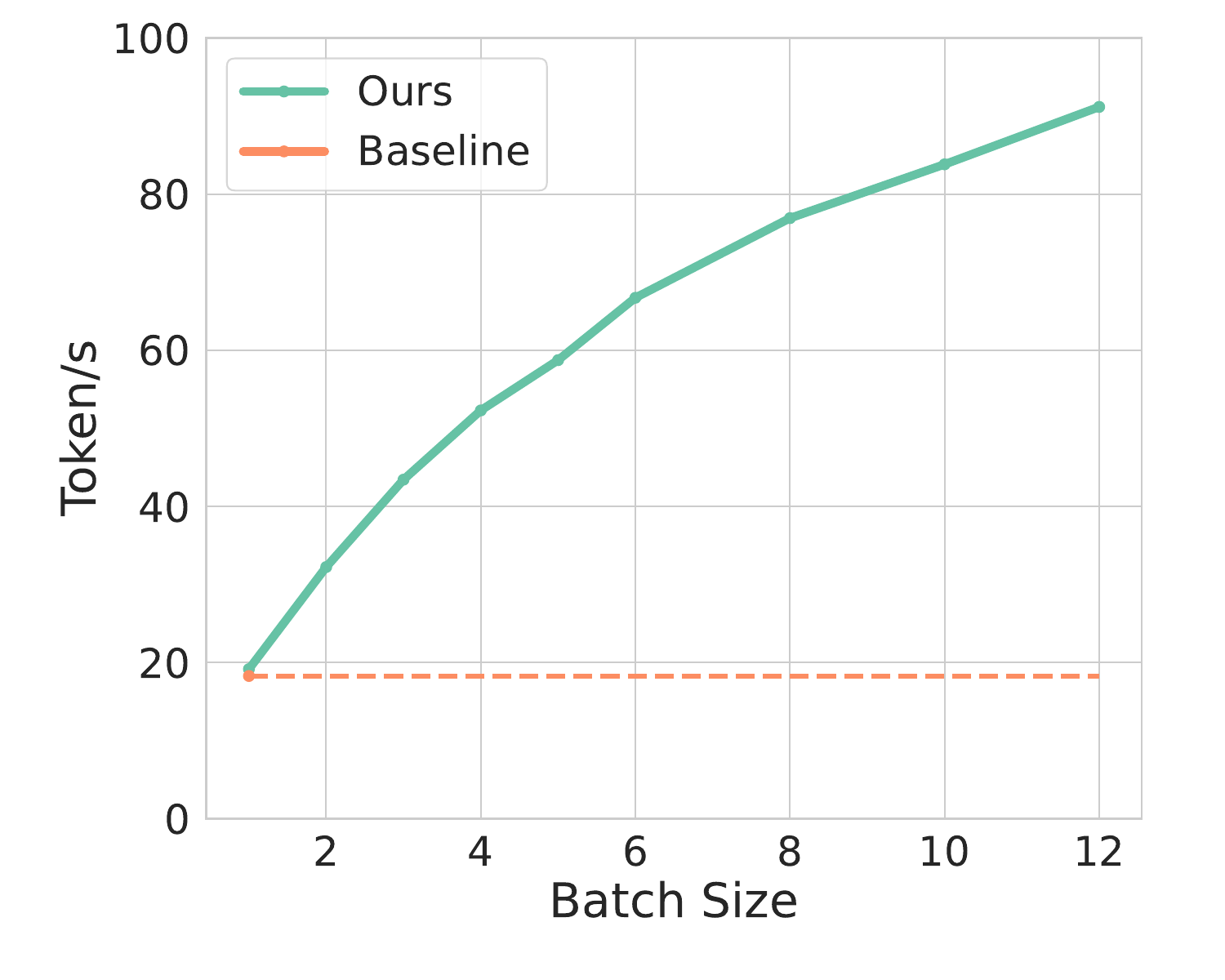}
    \label{fig: L40}
    }
  \subfloat[RTX 4090 with 8B model]{
\includegraphics[width=0.32\linewidth]{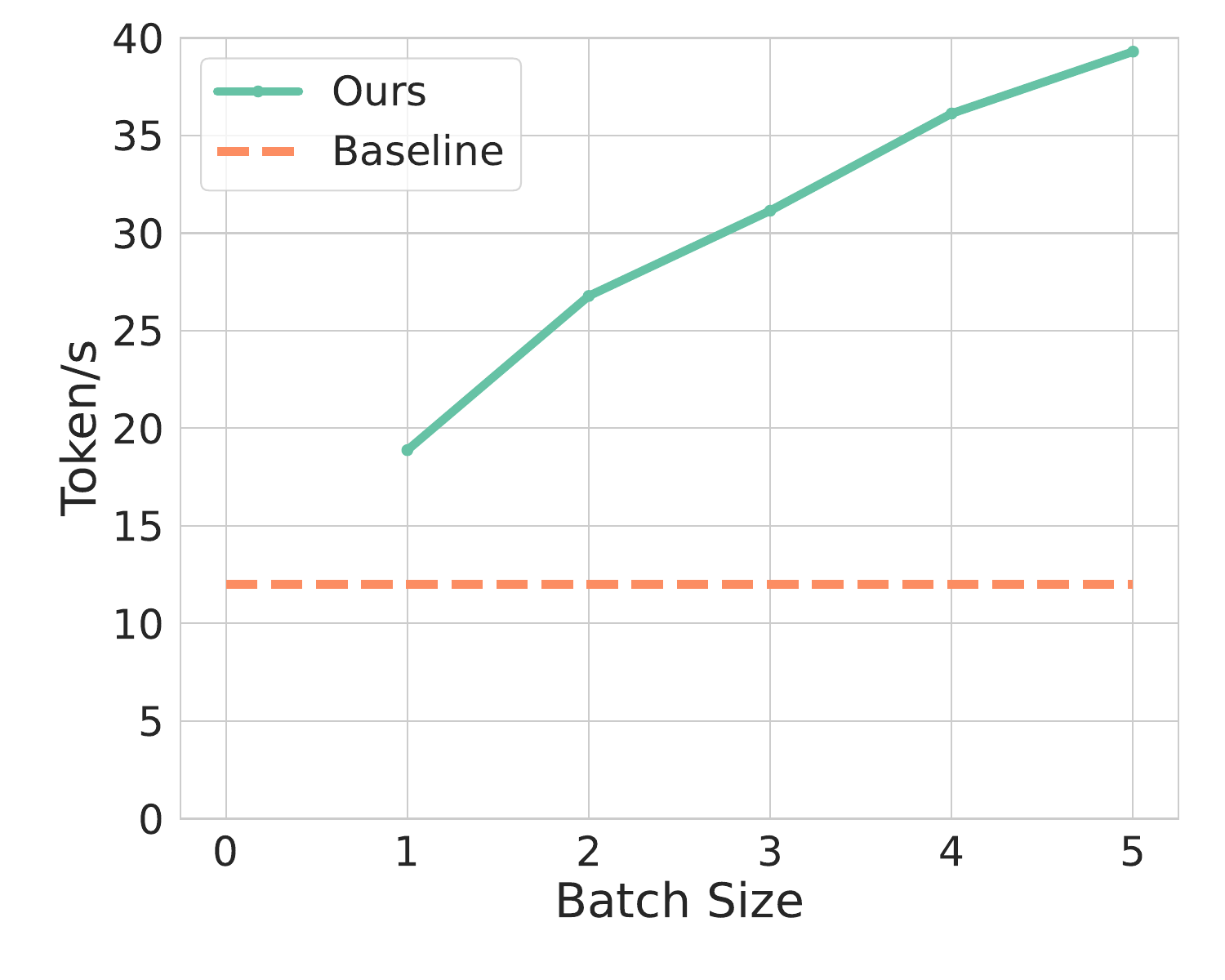}
    \label{fig: A5000} 
    }
    \caption{We evaluate \sys on three serving scenarios. \textbf{Left:} A100 serves 34B model with 16K context. \sys achieves $1.5\times$ throughput improvement.  \textbf{Mid:} L20 serves 13B model with 16K context. \sys achieves $5.0\times$ throughput improvement. \textbf{Right:} Simulated RTX 4090 serves 8B model with 96K context. \sys achieves a latency of 54ms in a single request serving and can improve the throughput of baseline by up to $3.3\times$. The dashed lines denote the highest throughput of baselines. With KV cache offloading, \sys can fit a much larger batch size compared with full attention on GPU, which contributes to the throughput improvement.}
    \vspace{-3mm}
\end{figure*}

\textbf{Setup.} We evaluate our system performance on 3 serving settings. (1) 80GB GPU (A100) and 34B model (CodeLlama-34B)~\citep{rozière2024codellamaopenfoundation} with 16K contexts; (2) 48GB GPU (L20) and 13B model (CodeLlama-13B)~\citep{rozière2024codellamaopenfoundation} with 16K contexts; (3) 24GB GPU\footnote{We simulate 24GB GPU by setting memory limit with L20. As the bandwidth of L20 (864GB/s) is less than  RTX 4090 (1TB/s), the real speed of our system should be slightly faster than the simulation.} (e.g. RTX 4090) and 8B model (Llama-3.1-8B)~\citep{dubey2024llama} with 96K contexts. 

\textbf{Baselines.} Our baselines for (1) and (2) are full attention on GPU, and for (3) is full attention on CPU with theoretical estimated bandwidth. Our system's GPU part is implemented in native Pytorch~\citep{paszke2019pytorch} and the CPU part in FBGEMM~\citep{fbgemm} in bfloat16 precision. Our CPU is Intel Platinum 8480+ for A100 and Intel  
 8563C for L20. In the last setting, the CPU bandwidth is estimated at 150GB/s, above the empirical bandwidth we measure when running a group query attention of size 4.

\textbf{Analysis.} In~\Cref{fig: A100,fig: A5000,fig: L40}, we demonstrate (1) \sys significantly improves decoding throughput for all three scenarios (A100: $1.5\times$, L20: $5.0\times$, RTX 4090: $3.3\times$) and can achieve a latency of 54ms for single request generation with 96K context for RTX 4090. (2) With KV cache offloading, \sys can fit much larger batches than GPU full attention baselines (over $12\times$). The ablation study of decoding throughput with different LSH hyper-parameters is presented in~\Cref{tab:throughput}.
\label{sec: efficiency}
\subsection{Ablation Study}
\label{sec: ablation}

\begin{figure*}
\vspace{-3mm}
    \centering
    \subfloat[Importance of centering]{
\includegraphics[width=0.32\linewidth]{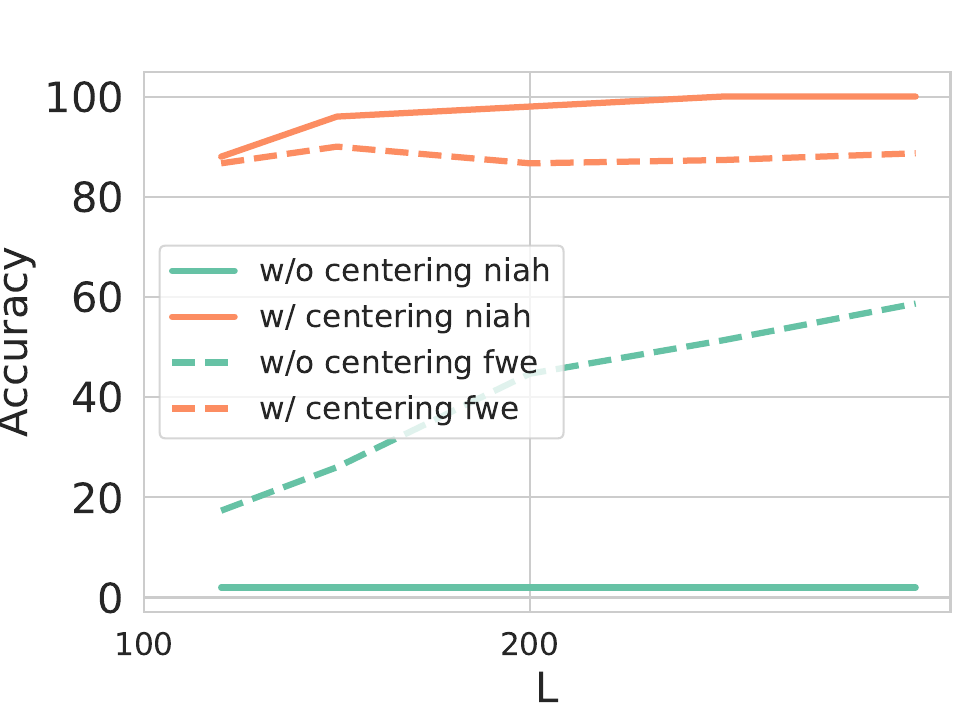}
    \label{fig: ablation centre}
    }
\subfloat[CWE]{
\includegraphics[width=0.32\linewidth]{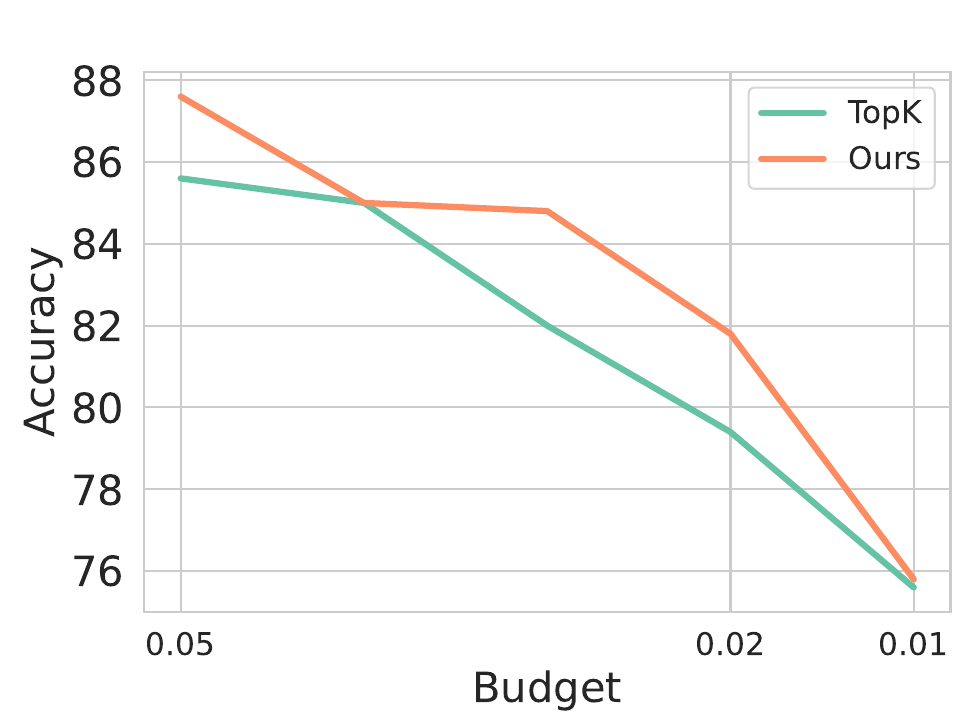}
    \label{fig: ablation cwe}
    }
  \subfloat[FWE]{
\includegraphics[width=0.32\linewidth]{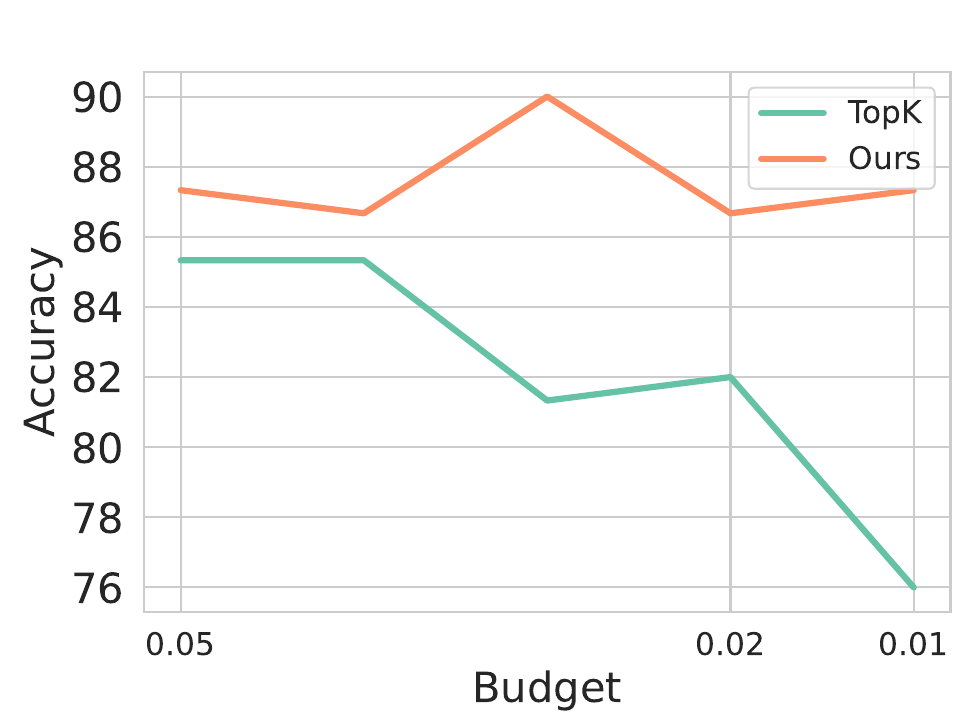}
    \label{fig: ablation fwe} 
    }
    \caption{\textbf{Left:} Accuracy comparison for with and without centering. Here we fix $K$ and vary $L$ for the two settings. \textbf{Mid and Right:} Comparison between \TopK attention and \sys. In the two aggregated tasks, sampling-based \sys can even beat the exact \TopK attention. The experiments are done on RULER~\citep{hsieh2024ruler} with a 16K context size.}
    \vspace{-4mm}
\end{figure*}

In this section, we empirically validate our two previous observations.  

\textbf{Centering is important for good performance.} In~\Cref{sec: approx via LSH}, we use a translation to center the keys before applying LSH sampling. Empirical results show this to be important for downstream tasks as shown in~\Cref{fig: ablation centre}. Without centering, the accuracy drops to almost zero in retrieval (NIAH) and degrades to $65\%$ in FWE.  We find almost no keys (less than $0.1\%$) can be sampled by the query without centering, as their orientation is almost opposite, as shown in~\Cref{fig: geometry}.

\textbf{Sampling goes beyond \TopK.} In~\Cref{fig: ablation cwe,fig: ablation fwe}, We compare the performance of \sys and \TopK attention in two aggregated tasks (CWE, FWE) where \TopK attention experiences significant performance degradation (\Cref{fig: topkperf}). \sys can even beat exact \TopK attention in these two tasks by a margin up to $3\%$ and $8\%$ respectively, demonstrating that sampling improves the ceiling of \TopK, which is impossible for a search-only algorithm.

\section{Conclusion}
In this work, we first present the limitation of \TopK attention approximation for addressing the computational and memory challenges of long-context LLM generation. 
Then we show oracle sampling can go beyond \TopK and introduce \sys, a novel approach that leverages LSH sampling to approximate the oracle sampling. 
\sys significantly reduces the workload of attention computation while preserving high accuracy across diverse tasks.
\sys relies on LSH sampling and a system co-design that offloads hash tables and reduced attention computation to the CPU. 
Our experimental results demonstrate that \sys substantially improves throughput and latency across multiple hardware configurations, outperforming traditional \TopK attention mechanisms. 
The theoretical soundness, robustness, and scalability of \sys open up new opportunities in both attention approximation methods and algorithm-hardware co-design.
\clearpage
\newpage
\bibliographystyle{assets/plainnat}
\bibliography{paper}

\clearpage
\newpage
\beginappendix
\section{Proofs for theorems}
\label{sec: proof}
\subsection{Proof for \Cref{theorem: unbias}}
\begin{proof}
\begin{align}
    \mathbb{E}(\Bar{o}) = \frac{1}{\mathcal{B}}\sum_{j=1}^{\mathcal{B}}\mathbb{E}[v_{i_j}] = \frac{1}{\mathcal{B}} \sum_{i=1}^{{n}}w_i v_i = o
\end{align}
Assume $\Sigma_1$ is the covariance matrix of $\Bar{o}$, $\Sigma_{2}$ is the covariance matrix of $v_i$
\begin{align}
    \text{Tr}(\Sigma_1) = \frac{1}{\mathcal{B}}\text{Tr}(\Sigma_2) = 
    \frac{1}{\mathcal{B}} (\mathbb{E}[||v_i||^2] - ||\mathbb{E}[v_i]||^2) 
    = \frac{1}{\mathcal{B}} (\mathbb{E}[||v_i||^2] - ||o||^2)
\end{align}
$\mathbb{E}[||v_{X}||^2] - ||o||^2$ is a constant, so the trace of covariance matrix monotonically decreases with $\mathcal{B}$.
\end{proof}

\subsection{Proof for \Cref{theorem: cost}}
\begin{proof}
\begin{align}
    \mathbb{E}[|S|] = \mathbb{E}\Big[\sum_{i=1}^{n}\textbf{1}_{i\in S}\Big] = \sum_{i=1}^{n}\mathbb{E}[\textbf{1}_{i\in S}] = \sum_{i=1}^{n}(1 - (1 - w_i)^{\mathcal{B}}) = n - \sum_{i=1}^{n}(1 - w_i)^{\mathcal{B}}
\end{align}
Without loss of generality, let $a_i = 1 - w_i$ and $a_1 = \min_{1\le i \le n}a_i = \eps$, then
\begin{align}
    \mathbb{E}[|S|] &= n - \sum_{i=1}^{n}a_i^{\mathcal{B}} = n - a_1^{\mathcal{B}}  - \sum_{i=2}^{n}a_i^{\mathcal{B}} \\
    &= n - \eps^{\mathcal{B}} - \sum_{i=2}^{n}a_i^{\mathcal{B}} 
\end{align}
$f(x) = x^{\mathcal{B}}$ is convex function with $\mathcal{B} \ge 1$ and $x \ge 0$. Then with Jensen's inequality, we have
\begin{align}
    \sum_{i=2}^{n}a_i^{\mathcal{B}}  &\ge 
    (n-1) \Big(\frac{\sum_{i=2}^{n}a_i}{n-1}\Big)^{\mathcal{B}} 
    = (n-1) \Big(\frac{(\sum_{i=1}^{n}a_i) - a_1}{n-1}\Big)^{\mathcal{B}} \\
    & = (n-1) (\frac{n - 1 - \eps }{n-1})^{\mathcal{B}} = (n-1) (1 - \frac{\eps }{n-1})^{\mathcal{B}} 
\end{align}
Let $g(x) = (1 - x)^\mathcal{B} +  \mathcal{B}x - 1$. We can prove $g(x) \ge 0 $ for any $x \in (0,1), \mathcal{B} \ge 1$.
Then we have 
\begin{align}
    \sum_{i=2}^{n}a_i^{\mathcal{B}} \ge (n-1) (1 - \frac{\eps \mathcal{B}}{n -1}) = n - 1 - \eps \mathcal{B}
\end{align}
Then we finally have
\begin{align}
    \mathbb{E}[|S|] = n - \eps^{\mathcal{B}} - \sum_{i=2}^{n}a_i^{\mathcal{B}}  \le 1 + \eps \mathcal{B}
\end{align}
\end{proof}

\section{Oracle sampling}
The optimal sampling probability to guarantee estimation is unbiased in terms of lowest variance is not directly using attention score distribution $w_i$, but $u_i' \propto w_i ||v_i||$. However, this sampling probability is not optimal in terms of downstream accuracy and efficiency. We attribute this to two reasons. First, we observe the value norm of the sink token is significantly smaller than others (\Cref{fig:vnorm}), given its lower probability of being sampled, which may influence the functionality of attention. Second, due to the same reason,  $u_i' \propto w_i ||v_i||$ is flatter than $w_i$, resulting larger computation cost (as analyzed by~\Cref{theorem: cost}).  
\label{sec: oracle}

\section{Supplementary analysis}

\begin{figure*}[hb]
    \centering
    {
\includegraphics[width=0.35\linewidth]{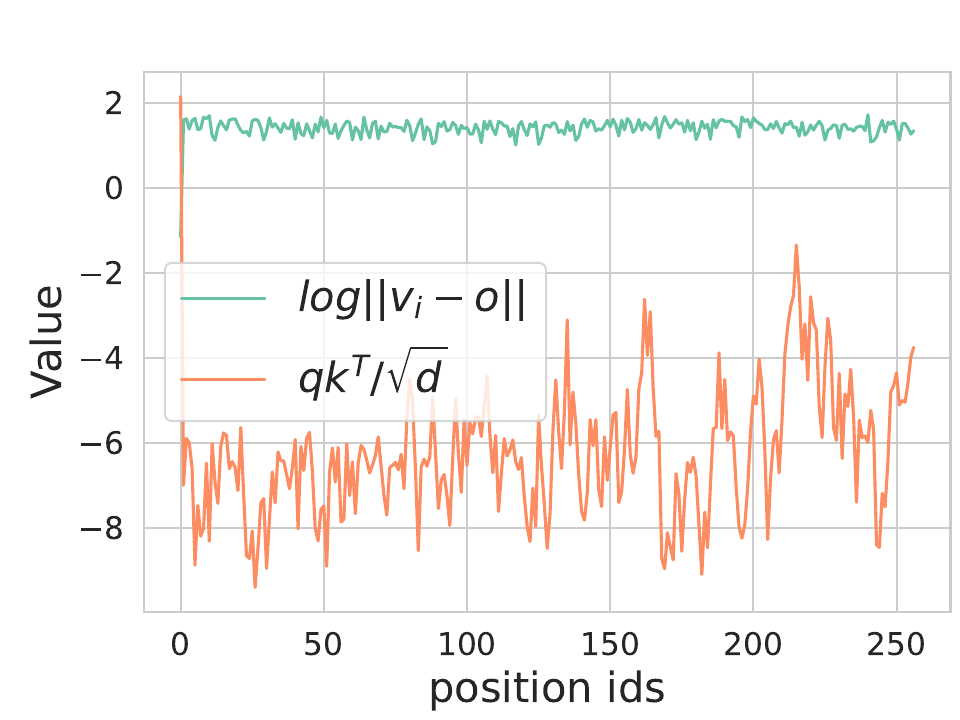}
    \label{fig: fluctuate}
    }
    \caption{The range of fluctuation of $\log|v_i - o|$ and $\frac{qk_i^T}{\sqrt{d}}$ in a single decoding step. Compared to $\frac{qk_i^T}{\sqrt{d}}$, $\log|v_i - o|$ is stable, hence we do not consider $\log|v_i - o|$ in our proposed sampling probability.}
    \label{fig:fluctuate}
\end{figure*}
\begin{figure*}[hb]
    \centering
    {
\includegraphics[width=0.35\linewidth]{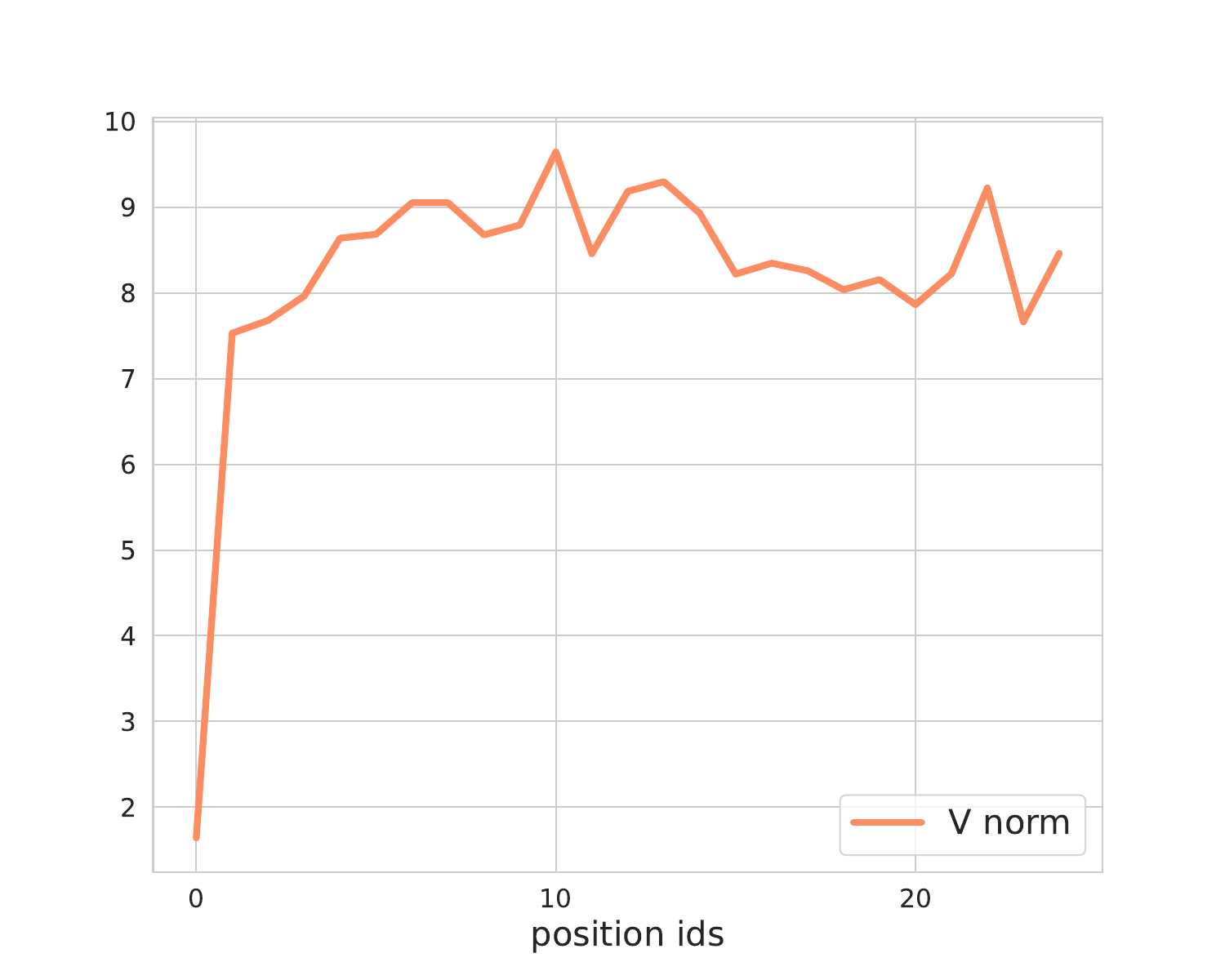}
    \label{fig: norm}
    }
    \caption{The $y$-axis is the norm of values states $\|v_i\|$ for token $i$ (on the x-axis). 
    We observe that the value norm $\|v_0\|$ of the attention sink is significantly smaller than others.}
    \label{fig:vnorm}
\end{figure*}

\Cref{fig:fluctuate} shows that compared to $\frac{qk_i^T}{\sqrt{d}}$, $\log|v_i - o|$ is stable in a decoding step. \Cref{fig:vnorm} shows that the norm of the value states of attention sink is smaller than others. 

\newpage
\section{Additional evaluation}
In this section, we provide additional experimental results to demonstrate that
\begin{itemize} [itemsep=0.0pt,topsep=0pt,leftmargin=*]
    \item \sys can support longer context lengths and a wide range of LLMs (\Cref{sec:ruler2}). 
    \item \sys can scale up with 70B level LLM (\Cref{sec: 70b ruler}).
    \item \sys can perform well in reasoning benchmarks (\Cref{sec: reasoning}). 
    \item \sys improves decoding throughput with various hyper-parameters (K, L). (\Cref{sec: system performance}).
\end{itemize}
\subsection{Longer Contexts}
\label{sec:ruler2}
Following the setups of~\Cref{tab:ruler}, we evaluate two additional models, MegaBeam-Mistral-7B-512K\footnote{\url{https://huggingface.co/aws-prototyping/MegaBeam-Mistral-7B-512k}} and Llama3-8B-Prolong-512K~\citep{gao2024train} with context lengths extended to 256K. The results are shown in~\Cref{tab:ruler2}.
\begin{table}[htbp]
\centering
\caption{Synthesized tasks on RULER~\citep{hsieh2024ruler}. \Sys preserves high accuracy with extended context lengths and different models. Config and cost are defined as in~\Cref{tab:lm-eval-harness}.}
\setlength{\tabcolsep}{3.2pt}
\small
\begin{tabular}{lc|rrrrrr|r|rr|r}
\toprule
Methods & Config & 16K & 32K & 64K & 96K & 128K & 256K & Avg. &$\text{Cost}_1$&$\text{Cost}_2$&$\text{Cost}_{total}$.\\ \midrule
\textit{MegaBeam-Mistral-7B-512K} & Full & 91.7&88.1 & 83.5& 83.7 & 83.5 & 82.5&85.5& 0.00& 1.00& 1.00\\
\rowcolor{pink!20} \Sys & (10,150) & 89.8&86.5 & 81.7&80.7& 81.6&79.0 &83.2 & 0.00& 0.02& 0.02\\
\rowcolor{pink!20} \Sys & (9,120) & 90.7&88.5 & 82.9&82.4&82.3&80.1& \textbf{84.5} & 0.00& 0.04& 0.04\\
\rowcolor{pink!20} \Sys & (8,75) & 90.6&86.4 & 82.8&81.6& 82.3&80.8&84.1 & 0.00& 0.05& 0.05\\
Quest & (16,0.04) & 83.3&83.2& 79.3&78.6& 78.5 & 78.5& 80.2 & 0.06& 0.04& 0.10\\
\midrule
\textit{Llama3-8B-Prolong-512K} & Full & 93.5&90.8 & 85.1& 83.5 & 81.7& 78.4 & 85.5& 0.00& 1.00& 1.00\\
\rowcolor{pink!20} \Sys & (10,150) & 88.0&86.4 & 81.3&78.8  &77.3&71.1 & 80.5 & 0.00& 0.02& 0.02\\
\rowcolor{pink!20} \Sys & (10,170) & 89.0&88.7 & 82.8&80.0  &77.7&73.7 & 82.0 & 0.00& 0.025& 0.025\\
\rowcolor{pink!20} \Sys & (9,120) & 91.4&88.2 & 82.4&80.4& 79.2 &75.2 & \textbf{82.8} & 0.00& 0.04& 0.04\\
\rowcolor{pink!20} \Sys & (8,75) & 91.4&88.6 & 83.1&80.5& 79.1& 73.9 & 82.8 & 0.00& 0.05& 0.05\\
Quest & (16,0.04) & 84.9&83.7& 78.7&78.6& 76.3 & 72.3 & 79.2 & 0.06& 0.04& 0.10\\
    \bottomrule
\end{tabular}
\label{tab:ruler2}
  \vspace{-3mm}
\end{table}

\subsection{Scaling up to larger models}
\label{sec: 70b ruler}
We evaluate \sys for meta-llama/Llama-3.1-70B-Instruct~\citep{dubey2024llama} to demonstrate that our approach can work well with larger LLMs in~\Cref{tab:70b ruler}. 

\begin{table}[htbp]
\centering
\caption{Synthesized tasks from RULER~\citep{hsieh2024ruler}. \Sys preserves high accuracy with low computation for 70B level models. 4 layers \{0,16,32,48\} are preserved.  Config and cost are defined as in~\Cref{tab:lm-eval-harness}.}
  \vspace{-2mm}
\setlength{\tabcolsep}{3.5pt}
\small
\begin{tabular}{lc|rrrr|r|rr|r}
\toprule
Methods & Config & 16K & 32K & 64K & 96K & Avg. &$\text{Cost}_1$&$\text{Cost}_2$&$\text{Cost}_{total}$.\\ \midrule
\textit{Llama-3.1-70B-Instruct} & Full & 96.4&94.6 & 89.2& 80.8 & 90.3& 0.00& 1.00& 1.00\\
\rowcolor{pink!20} \Sys & (10,150) & 94.7&93.5 & 87.5& 79.3 &88.8 & 0.00& 0.02& 0.02\\
\rowcolor{pink!20} \Sys & (9,110) & 95.7&93.5 & 88.4& 79.4 &89.3 & 0.00& 0.034& 0.034\\
\rowcolor{pink!20} \Sys & (9,120) & 95.5&94.1 & 88.8& 80.6&89.8& 0.00& 0.04& 0.04\\
\bottomrule
\end{tabular}
\label{tab:70b ruler}
  \vspace{-3mm}
\end{table}

\subsection{Reasoning}
\label{sec: reasoning}
In mathematical reasoning tasks infini\_igsm~\citep{IGSM4K,IGSM8K}, \sys consistently outperforms Quest \citep{tang2024quest} across all complexity (in terms of operators). We also find \TopK attention suffers from significant performance degradation while Oracle Sampling can maintain high accuracy. 
\begin{table}[htbp]
\centering
\caption{Tasks from infini\_igsm~\citep{IGSM4K,IGSM8K}. \Sys preserves high accuracy for reasoning tasks. Config and cost for \sys and Quest are defined as in~\Cref{tab:lm-eval-harness}. Config denotes the ratio of selected tokens for \TopK and sampled tokens for oracle sampling. For oracle sampling, massive duplication exists in sampled tokens, so $\text{Cost}_2$ is significantly lower than the ratio of sampled tokens \Cref{theorem: cost}.}
  \vspace{-2mm}
\setlength{\tabcolsep}{3.5pt}
\small
\begin{tabular}{c|lc|rrr|rr|r}
\toprule
Task & Methods & Config & 2-Ops & 4-Ops & 5-Ops  &$\text{Cost}_1$&$\text{Cost}_2$&$\text{Cost}_{total}$.\\ \midrule
\multirow{11}{*}{\textit{4K close}~\citep{IGSM4K}} &
\textit{Llama-3.1-8B-Instruct} & Full & 87.4&71.4 & 26.8&   0.00& 1.00& 1.00\\
\rowcolor{pink!20} &  \Sys & (10,300) & 83.1&67.2 & 20.7&  0.00& 0.06& 0.06\\
\rowcolor{pink!20} &  \Sys & (10,220) & 79.8&58.9 & 17.9& 0.00& 0.04& 0.04\\
 \rowcolor{pink!20} & \Sys & (10,150) & 68.3&43.5 & 11.7&  0.00& 0.02& 0.02\\
 & \TopK & 0.06 & 78.6&62.9 & 20.8 & 0.50& 0.06& 0.56\\
 & \TopK & 0.04 & 76.2&59.0 & 19.2 & 0.50& 0.04& 0.54\\
 & \TopK & 0.02 & 71.5&44.0 & 11.3 & 0.50& 0.02& 0.52\\

\rowcolor{orange!10} & Oracle Sampling & 0.3 & 88.1&72.4 & 27.6 & 0.50& 0.02& 0.52\\
\rowcolor{orange!10} & Oracle Sampling & 0.1 & 88.5&69.2 & 26.2 & 0.50& 0.01& 0.51\\
\rowcolor{orange!10} & Oracle Sampling & 0.02 & 83.1&57.9 & 11.9 & 0.50& 0.005& 0.505\\
& Quest & (16,0.06) & 55.8&23.2 & 5.2 & 0.06& 0.06& 0.12\\
\midrule
\multirow{11}{*}{\textit{8K close}~\citep{IGSM8K}} &
\textit{Llama-3.1-8B-Instruct} & Full & 80.2&68.8 & 26.0&   0.00& 1.00& 1.00\\
\rowcolor{pink!20} &  \Sys & (10,300) & 78.6&61.5 & 25.2&  0.00& 0.06& 0.06\\
\rowcolor{pink!20} &  \Sys & (10,220) & 72.2&60.7 & 20.4&  0.00& 0.04& 0.04\\
 \rowcolor{pink!20} & \Sys & (10,150) & 67.1&44.0 & 11.9 & 0.00& 0.02& 0.02\\
 & \TopK & 0.06 & 70.2&61.1 & 22.3 & 0.50& 0.06& 0.56\\
 & \TopK & 0.04 & 66.9&55.2 & 20.6 & 0.50& 0.04& 0.54\\
 & \TopK & 0.02 & 64.7&47.2 & 15.9 & 0.50& 0.02& 0.52\\

\rowcolor{orange!10} & Oracle Sampling & 0.3 & 80.0&67.3 & 26.2 & 0.50& 0.02& 0.52\\
\rowcolor{orange!10} & Oracle Sampling & 0.1 & 76.6&64.1 & 25.4 & 0.50& 0.01& 0.51\\
\rowcolor{orange!10} & Oracle Sampling & 0.02 & 79.0&60.3 & 20.4 & 0.50& 0.005& 0.505\\
& Quest & (16,0.06) & 54.8&30.0 & 11.1 & 0.06& 0.06& 0.12\\
\bottomrule
\end{tabular}
\label{tab:reasoning}
  \vspace{-3mm}
\end{table}

\subsection{System performance}
\label{sec: system performance}
In this section, we evaluate the system performance (latency, throughput) of \sys under different hyper-parameter configurations. We use Llama-3.1-8B-Instruct~\citep{dubey2024llama} with 96K contexts as an example. 

\begin{table}
\centering
\setlength{\tabcolsep}{3.5pt}
\small
\caption{System performance for \sys using Llama-3.1-8B-Instruct with a 96K context length under varying hyper-parameter configurations. We report the decoding latency (time between tokens, TBT) when the batch size is 1, the maximum throughput, and the throughput with a latency constraint of 200ms ($\text{Throughput}_{\text{200ms}}$ in the table). Config and cost are defined as in~\Cref{tab:lm-eval-harness}. The number with $^*$ means hit the memory limit of CPU.}
  \vspace{-2mm}
\setlength{\tabcolsep}{3.5pt}
\begin{tabular}{c|ccc|r}
\toprule
 Config & TBT (ms) & Max Throughput (tokens/sec) & $\text{Throughput}_{\text{200ms}}$ (tokens/sec) &$\text{Cost}_{total}$.\\ 
 \midrule
 (11,300) & 17.38&41.68$^*$ & 40.84& 0.02\\
 (10,220) & 14.07&32.29$^*$ & 26.66& 0.04\\
 (10,170) & 16.79&46.52$^*$ & 39.90& 0.025\\
 (10,150) & 18.31&53.78 & 48.89& 0.02\\
 (9,120) & 13.93&32.50 & 26.60& 0.04\\
 (8,75) & 12.47&27.43 & 21.17& 0.05\\
\bottomrule
\end{tabular}
\label{tab:throughput}
  \vspace{-3mm}
\end{table}

\section{Selection of hyper-parameter (K, L)}
\label{sec: hyper-parameter selection}
In this section, we discuss the impact of the LSH hyper-parameter (K, L) and how to select it. First, we briefly explain what hyper-parameter (K, L) does for LSH sampling. Then, we explain the relations between (K, L) and attention computation cost and accuracy. Finally, we show how we decide the parameters by ablation studies.  

\subsection{(K, L) in LSH}
 In each hash table, we use K hash functions to compute the hash code of $k$ and $q$.  In Simhash~\citep{10.1145/509907.509965}, the hashing we use in \sys, the hash functions are random projections. With K random projections, we are able to partition the space (in our problem, the space is $R^{128}$) into $2^K$ subspace. If and only if $k$ and $q$ fall in the same subspace, we say they collide in this hash table.  We have L hash tables in total. In \sys, if and only if $k$ and $q$ collide in at least two hash tables, $k$ is sampled by $q$.  Here are some intuitions about how (K, L) will influence the LSH sampling in \sys. 
 \begin{itemize} [itemsep=0.0pt,topsep=0pt,leftmargin=*]
    \item If K is too small, then we cannot partition the space well; we will sample too many $k$s, which might be far away from $q$ (in the attention problem, this means their inner production is small), increasing computation cost.
    \item On the other hand, if K is too large,  although the quality of sampled ks will be better, the collision probability in each table will be small; thus, the number of the sampled ks will be reduced. We need to increase L to ensure that a certain number of keys are sampled and involved in the computation.  However, increasing (K, L) too much will bring more memory overhead on CPU DRAM since we build L hash tables for each key-value head. 
\end{itemize}

Thus, (K, L) is important because it balances computation cost, overhead, and sampling quality (which determines accuracy). Tuning (K, L) is necessary in LSH~\citep{10.14778/3137765.3137836,6242372}.

\subsection{(K, L) and memory overhead}
(K, L) will change two overheads brought by \sys: the memory occupied by hash tables on the CPU and extra computation for random projections (hash functions) on the GPU (as shown in~\Cref{tab: lsh overhead}). 

\begin{table}[htbp]
\centering
\setlength{\tabcolsep}{3.5pt}
\small
\caption{The overhead of Locality sensitive hashing during decoding. We report the size of random projectors (on GPU) and hash tables (on CPU), the computation overhead \textbf{CO} (refers to the ratio between computation introduced by random projections in LSH and the computation of the original model's linear projections (e.g., $W_Q, W_K, W_V$, and MLP)).  Notice that when the context length exceeds 64K, we need to use 32-bit integers to store the indices for the KV cache in hash tables. Llama-3.1-8B/70B-Instruct~\citep{dubey2024llama} and Code-Llama-34b-16K~\cite{rozière2024codellamaopenfoundation} use group query attention, thus the sizes of hash tables are reduced.} 
\begin{tabular}{lcc|cc|c}
\toprule
Models & (K, L) & Context length & Projectors & Hash tables & CO  \\ 
\midrule
\textit{Llama-3.1-8B-Instruct} & (10, 150) & 96K & 384KB & 14GB & $3.8\%$\\
\textit{Llama-3.1-8B-Instruct} & (11, 300) & 96K & 825KB & 28GB & $8.5\%$\\
\textit{Llama-3.1-8B-Instruct} & (10, 150) & 64K & 384KB & 4.7GB & $3.8\%$\\
\textit{Llama-3.1-70B-Instruct} & (10, 150) &64K & 384KB & 11.8GB & $1.8\%$\\
\textit{Code-Llama-13b-16K} & (10, 150) & 16K & 384KB & 7.3GB & $5.2\%$\\
\textit{Code-Llama-34b-16K} & (10, 150) & 16K & 384KB & 1.8GB & $2.2\%$\\
\bottomrule
\end{tabular}
\label{tab: lsh overhead}
  \vspace{-2mm}
\end{table}

LLM decoding is a memory-bandwidth-bound process and the majority of time is spent loading the data (parameters/KV cache) to GPU cores rather than actually doing the computation~\citep{miao2023specinfer,self_spec,chen2024sequoia}. Besides, the time-consuming part, i.e., the long-context attention computation, is moved to the CPU. Thus, the $1.8\% \sim 8.5\%$ extra computation on GPU will only make a minor difference in execution time. However, the enlarged size of hash tables prevents us from always increasing (K, L) to get more accurate results.

As shown in~\Cref{tab: lsh overhead}, under the same (K, L), the memory overhead of hash tables grows linearly with context length and the total number of key-value heads in models (which is determined by model sizes). 

\subsection{(K, L) and computation cost/budget} In summary, increasing K will make the budget\footnote{$\text{Cost}_2$ in~\Cref{tab:lm-eval-harness,tab:ruler,tab:longbench}} smaller, and increasing L will increase the budget. 

Theoretically, as introduced in~\Cref{sec: approx via LSH}, in our approach, the key $k_i$ is sampled only if at least two hash tables exist where $k_i$ shares the hash value with query q. With the assumption that $k_i$ is well-distributed (In each hash table out of L, each hash value corresponds to roughly the same number of $k_i$s), the ratio of retrieved $k_i$s can be estimated with
\begin{align}
    \mathcal{B} / n = 1 - (1 - 0.5^{K})^{L} - L \times 0.5^K (1 - 0.5^K)^{L-1} 
    \label{eq: budget}
\end{align}
 where $n$ is the context length. Here,  we estimate the collision probability of $k_i$ and $q$ in a single hash table as $0.5^K$. 

 Empirically, the ratio of retrieved keys and values ($\mathcal{B} / n$) might differ from the above estimation since the data is not perfectly distributed.  We present the empirically measured budget in~\Cref{tab: budget}.
\begin{table}[htbp]
\centering
\small
\caption{Empirical measured budget/cost for different (K, L).}
\resizebox{0.5\linewidth}{!}{
\begin{tabular}{l|cccccc}
\toprule
 K / L & 75 & 100 & 120 & 150 & 200 & 300 \\ 
 \midrule
 7 &  $14\%$ & $21\%$ & $27\%$ & $35\%$ & $48\%$ & $66\%$\\
 8 &  $5\%$ & $8\%$ & $11\%$ & $15\%$ & $22\%$ &$36\%$ \\
 9 &  $1.6\%$ & $2.7\%$ & $4\%$ & $5.4\%$ & $8.5\%$ &$15.4\%$ \\
 10 & $0.5\%$ & $0.9\%$ & $1.5\%$ & $2\%$ & $3\%$ &$6\%$ \\
 11 & $0.15\%$ & $0.3\%$ & $0.5\%$ & $0.6\%$ & $1\%$ &$2\%$ \\
\bottomrule
\end{tabular}
}
\label{tab: budget}
  \vspace{-2mm}
\end{table}

\subsection{(K, L) and accuracy}
There are no naive relations between (K, L) and downstream accuracies since (K, L) not only influences sampling quality but also the computation budget. One safe way to discuss the relation between (K, L) and accuracy is: Fixing the computation budget, larger (K, L) will potentially produce higher accuracy, since the sampling quality is higher. Our experimental results show that, 
\begin{itemize} [itemsep=0.0pt,topsep=0pt,leftmargin=*]
\item Increasing (K, L) can significantly improve accuracy in relatively longer contexts~\Cref{tab: large table}.
\begin{table}[htbp]
\centering
\small
\caption{We show the effectiveness of larger hash tables for longer contexts by evaluating MegaBeam-Mistral-7B-512K on RULER~\citep{hsieh2024ruler}. With the same computation cost ($\sim2\%$), config (11, 300) achieves higher accuracy compared to (10, 150).}
\begin{tabular}{l|ccc}
\toprule
 (K, L) & 16K & 128K & 256K \\ 
 \midrule
 Full & 91.7 & 83.7 & 82.5 \\
 (10, 150) & 89.8 & 80.7& 79.0 \\
 (11, 300) & 90.6& 83.3 & 81.9 \\
\bottomrule
\end{tabular}
\label{tab: large table}
  \vspace{-2mm}
\end{table}
\item Same set of (K, L) can generalize to larger LLMs~\Cref{tab: scale}.
\begin{table}[htbp]
\centering
\small
\caption{8B and 70B models on RULER~\citep{hsieh2024ruler} 64K.}
\begin{tabular}{l|ccccc}
\toprule
 Models/Config & Full & (10, 150) & (10, 135) & (9, 120) & (9, 110)\\ 
 \midrule
 Llama-3.1-8B-Instruct & 86.1 & 84.8 & 83.6  & 84.7 &  84.7 \\
 Llama-3.1-70B-Instruct & 89.2 & 87.5 & 86.7 & 88.8 & 88.4 \\
\bottomrule
\end{tabular}
\label{tab: scale}
  \vspace{-2mm}
\end{table}
\end{itemize}

\subsection{How to select (K, L)}
Finding the optimal (K, L) for high accuracy as well as efficiency is a long-standing problem in LSH. Similar to the traditional hyper-parameter tuning process in machine learning, K, and L are configured offline based on data subsets. In LSH, K is a more sensitive hyper-parameter than L.  A slight change of K can drastically influence the number of retrieved items (i.e., budget/cost) and quality. In \sys, K=8-10 is manually determined by ablations on small-scale tasks and found to be effective across various models and tasks; then, we adjust L to obtain the wanted computation cost/budget.

Here, we present two ablations to demonstrate the selection of K in~\Cref{tab: abk1,tab: ablationk}.
\begin{table}[htbp]
\centering
\small
\caption{Fixing the budget/cost to $4\%$, we ablation the performance of different (K, L) on RULER~\citep{hsieh2024ruler} 16K.}
\begin{tabular}{l|ccccc}
\toprule
 Models/Config & Full & (10, 240) & (9, 120) & (8, 65) & (7, 35)\\ 
 \midrule
 Llama-3.1-8B-Instruct & 94.2 & 94.2 & 92.8  & 92.3 &  88.5 \\
\bottomrule
\end{tabular}
\label{tab: abk1}
  \vspace{-2mm}
\end{table}

\begin{table}[htbp]
\centering
\small
\caption{Fixing L as 120, we ablation the performance of different K on RULER~\citep{hsieh2024ruler} 16K for Llama-3.1-8B-Instruct.}
\begin{tabular}{l|ccccc}
\toprule
 (K, L) & Full & (10, 120) & (9, 120) & (8, 120) & (7, 120)\\ 
 \midrule
 Cost & 1.0 & 0.012 & 0.04  & 0.11 & 0.27 \\
 Accuracy & 94.2 & 92.8 & 92.8 & 94.1 & 94.3 \\
\bottomrule
\end{tabular}
\label{tab: ablationk}
  \vspace{-2mm}
\end{table}

 If we want the computation cost to be below $5\%$ and L below 200 (to reduce memory overhead in the CPU), then K=8-10 is a reasonable choice. Unlike K, L is not that sensitive. We select L based on the following principle after determining K: we can allow the computation cost to be smaller for larger K since the sampling is more precise. This is why we choose to use (8, 75), (9, 120), and (10, 150).

 It’s worth pointing out that tuning (K, L) is a challenging and long-standing problem in LSH, and we only give an example of practice in \sys. More advanced hashing algorithms (such as Cross-polytope~\citep{andoni2015practical} or data-dependent ones~\citep{andoni2015optimal}) can improve the trade-off between memory overhead and accuracy. We leave it as a future direction. 

\section{TopK vs. Sampling}
\label{sec:sampling example}
In this section, we provide an intuitive understanding of how sampling can work better than TopK. TopK only captures the ranking information when estimating attention output. In contrast, sampling considers the entire data distribution (i.e., the attention score after $\text{Softmax}$). 

Here is an example. Imagine a zoo with 100 animals: 10 elephants, 10 pigs, 10 tigers, and 70 other unique animals. The daily food consumption for each group is as follows:
\begin{itemize}
    \item \textbf{Elephants}: 50 lb/day each
    \item \textbf{Pigs}: 20 lb/day each
    \item \textbf{Tigers}: 10 lb/day each
    \item \textbf{Other unique animals}: 1 lb/day each
\end{itemize}
To compute the true average daily food consumption per animal in the zoo:
\[
\text{True Average} = \frac{(10 \times 50) + (10 \times 20) + (10 \times 10) + (70 \times 1)}{100} = 8.7 \, \text{lb}.
\]

If we use a \textbf{Top-K approach} (e.g., selecting the top 10 animals based on the numbers of animals), we include elephants, pigs, tigers, and 7 randomly selected animals from the unique ones. The estimated average is:
\[
\text{TopK Average} = \frac{(10 \times 50) + (10 \times 20) + (10 \times 10) + (7 \times 1)}{37} = 22 \, \text{lb}.
\]
This overestimates the average because it disproportionately weights high-consumption animals.

Instead, we perform \textbf{sampling with replacement} from the animal distribution, proportional to their numbers. The probabilities for each group are:
\[
\text{Sampling Probabilities} = [0.1, 0.1, 0.1, 0.01 \times 70],
\]
where 0.1 represents the probabilities for elephants, pigs, and tigers (10/100 each), and 0.01 corresponds to each unique animal (1/100).

Perform 10 random draws. A possible sampling outcome could be: \texttt{[elephant, pig, tiger, other, other, other, other, other, other, other]}. The corresponding daily food estimate is:
\[
\text{Sample Estimate} = \frac{50 + 20 + 10 + (7 \times 1)}{10} = 8.7 \, \text{lb}.
\]
This estimate is unbiased, meaning the expected value of the estimates equals the true average (8.7 lb). While there is variance across individual trials, the \textbf{standard deviation (std)} can be calculated as 4.7 lb for a 10-sample budget.

Increasing the sampling budget reduces variance. For example, with 20 samples, the \textbf{std} decreases to 3.4 lb. Meanwhile, Top-K with a budget of 20 adds 17 unique animals, yielding:
\[
\text{TopK Average (K=20)} = \frac{(10 \times 50) + (10 \times 20) + (10 \times 10) + (17 \times 1)}{47} = 17 \, \text{lb}.
\]
Again, the Top-K estimate remains biased, significantly overestimating the average.

Note that this is intended as an intuitive example. For a detailed and formal derivation of the sampling methodology, please refer to \citet{kloek1978bayesian,mcbook,lohr2021sampling}.

\section{Limitations and future work}
\sys stores the offloaded KV cache and hash tables on CPU DRAM, which is unsuitable for serving scenarios with insufficient DRAM. KV cache quantization methods like QServe~\citep{lin2024qserve} and KIVI~\citep{liu2024kivi} can help to reduce the KV cache memory. Currently, another limitation is that, we have not implemented \sys in prefilling stage, which is also an important direction in long context LLM serving.  Applying more advanced LSH algorithms, such as Cross-polytope hash~\citep{andoni2015practical}, can reduce the size of hash tables while improving estimation accuracy. Building CPU-GPU pipelines~\citep{he2024fastdecode} and leveraging the new avx512\_bf16 features of CPUs will improve efficiency. For higher-end GPUs with sufficient HBM,  leveraging LSH to accelerate GPU attention computation is also an interesting topic, as GPU-friendly LSH algorithms and efficient GPU kernels~\citep{app10072539,9635657} are required to do sampling. Besides, how to automatically tune the LSH hyper-parameter (K, L)~\citep{10.14778/3137765.3137836} is also an interesting future work.

\end{document}